\documentclass[sigconf]{acmart}

\usepackage{subcaption}
\usepackage{graphicx}
\usepackage{enumitem}
\usepackage{multirow}
\usepackage{soul}
\usepackage{rotating}
\usepackage{stfloats}
\usepackage{makecell}
\usepackage[ruled,boxed,linesnumbered]{algorithm2e}
\usepackage{tabularx}
\usepackage{colortbl}
\usepackage{bbding}
\usepackage{xcolor}
\usepackage{bigstrut}
\usepackage{threeparttable}
\usepackage{diagbox}
\usepackage{fontawesome5}
\definecolor{mygray}{rgb}{0.92,0.97,0.92}
\definecolor{light}{gray}{0.92}
\newcounter{problemname}
\newcounter{definitionname}

\newtheorem{definition}{Definition}
\newcommand{\model}{\textsc{Proceed} }
\newcommand{\modelns}{\textsc{Proceed}}
\definecolor{cadmiumgreen}{rgb}{0.0, 0.42, 0.24}
\newcommand{\icon}{{\textcolor{cadmiumgreen}{\faPlayCircle[regular] }}}
\newcommand{\eat}[1]{}

\graphicspath{ {./figures/} }

\AtBeginDocument{%
  }
\copyrightyear{2025}
\acmYear{2025}
\setcopyright{acmlicensed}
\acmConference[KDD '25] {Proceedings of the 31st ACM SIGKDD Conference on Knowledge Discovery and Data Mining V.1}{August 3--7, 2025}{Toronto, ON, Canada.}
\acmBooktitle{Proceedings of the 31st ACM SIGKDD Conference on Knowledge Discovery and Data Mining V.1 (KDD '25), August 3--7, 2025, Toronto, ON, Canada}
\acmISBN{979-8-4007-1245-6/25/08}
\acmDOI{10.1145/3690624.3709210
}
\begin{document}
\title{Proactive Model Adaptation Against Concept Drift for Online Time Series Forecasting}

\author{Lifan Zhao}
\affiliation{%
\institution{Shanghai Jiao Tong University}
\country{Shanghai, China}
}
\email{mogician233@sjtu.edu.cn}

\author{Yanyan Shen}
\affiliation{%
  \institution{Shanghai Jiao Tong University}
  \country{Shanghai, China}
  }
\email{shenyy@sjtu.edu.cn}

\renewcommand{\shortauthors}{Lifan Zhao \& Yanyan Shen}

\begin{abstract}
Time series forecasting always faces the challenge of concept drift, where data distributions evolve over time, leading to a decline in forecast model performance. Existing solutions are based on online learning, which continually organize recent time series observations as new training samples and update model parameters according to the forecasting feedback on recent data. However, they overlook a critical issue: obtaining ground-truth future values of each sample should be delayed until after the forecast horizon. This delay creates a temporal gap between the training samples and the test sample. Our empirical analysis reveals that the gap can introduce concept drift, causing forecast models to adapt to outdated concepts. In this paper, we present \textsc{Proceed}, a novel proactive model adaptation framework for online time series forecasting. \textsc{Proceed} first estimates the concept drift between the recently used training samples and the current test sample. It then employs an adaptation generator to efficiently translate the estimated drift into parameter adjustments, proactively adapting the model to the test sample. To enhance the generalization capability of the framework, \textsc{Proceed} is trained on synthetic diverse concept drifts. Extensive experiments on five real-world datasets across various forecast models demonstrate that \textsc{Proceed} brings more performance improvements than the state-of-the-art online learning methods, significantly facilitating forecast models' resilience against concept drifts. Code is available at \url{https://github.com/SJTU-DMTai/OnlineTSF}.

\end{abstract}

\ccsdesc[500]{Information systems~Data stream mining}
\ccsdesc[100]{Computing methodologies~Online learning settings}

\keywords{Time series forecasting, Online learning, Concept drift}


\maketitle

\section{Introduction}
Time series forecasting has been a prevalent task in numerous fields such as climate~\cite{Pangu}, energy~\cite{SolarNIPS, Sun2023DSE}, retail~\cite{Boese2017},  and finance~\cite{DoubleAdapt}. Recent years have witnessed a surge of deep learning-based forecast models~\cite{PatchTST, iTransformer, GPT4TS, MOIRAI, SOFTS} that take past time series observations to predict values in the next $H$ steps, where $H$ is called \textit{forecast horizon}.
Due to the dynamic nature of the environment, latent \textit{concepts} that influence observation values (e.g., social interest~\cite{DRAIN}, stock market sentiment~\cite{DDGDA}) often change over time. This ubiquitous phenomenon is known as \textit{concept drift}~\cite{SurveyConcepDrift}.  In the presence of concept drift, future test data may not follow a similar data distribution as the historical training data, causing degradation in forecasting performance~\cite{LLF, DoubleAdapt, duan2025brain}. 

\textit{Online learning} is commonly used to mitigate the effects of concept drift. The key consideration is to 
continually transform the newly observed time series into a set of training samples and adjust model parameters by minimizing the forecast errors on the new training samples. 
In addition to the standard fine-tuning technique, recent works~\cite{FSNet, OneNet} proposed more advanced model adaptation techniques, which focus on how to effectively adapt to recent data by utilizing \textit{forecasting feedback} (e.g., errors or gradients) on the new training samples. Among them, FSNet~\cite{FSNet} monitors the gradients in previous fine-tuning, transforms them into parameter adjustments, and tailors a new forecast model to the current training samples.
OneNet~\cite{OneNet} is an online ensembling network that generates ensemble weights to combine two forecast models and dynamically adjusts the ensemble weights and the models' parameter weights according to the forecast errors. 
However, 
existing online learning methods overlook the fact that
the ground truth of each prediction is not available immediately but is delayed after the forecast horizon.

\begin{figure}[t]
  \centering
  \includegraphics[width=\linewidth]{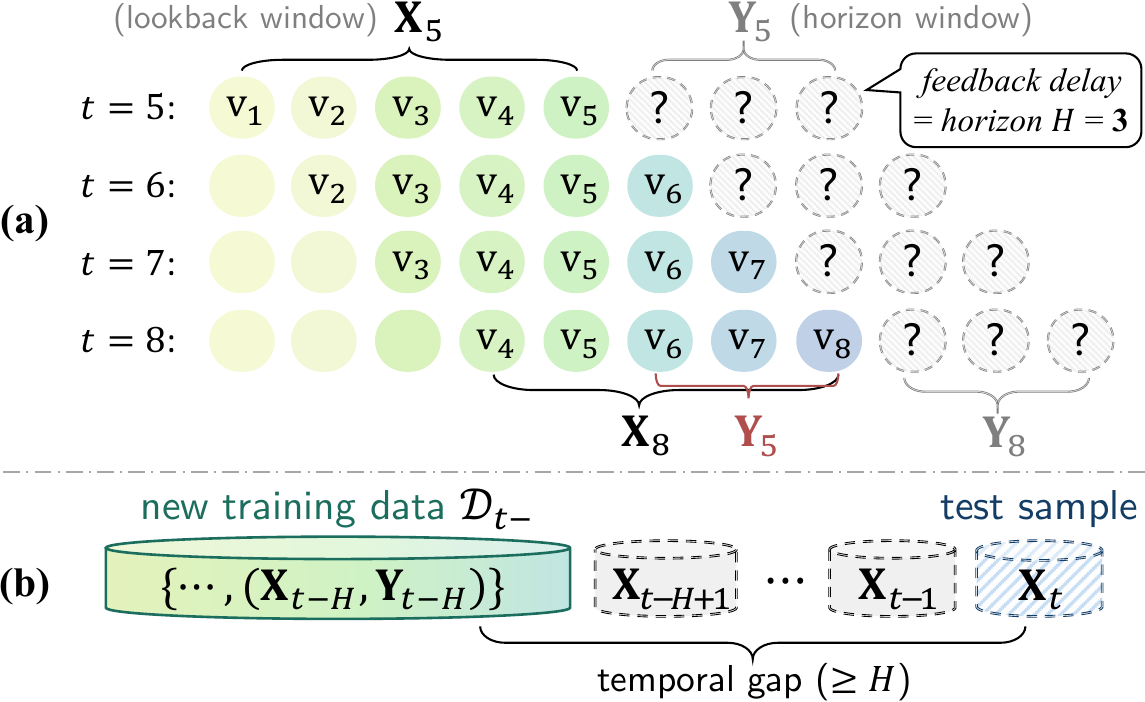}
  \caption{(a) Example of feedback delay when $H=3$. At online time $t=5$, we use observation values $\mathbf X_5=\{\mathbf v_1,\cdots,\mathbf v_5\}$ to forecast future values $\mathbf Y_5=\{\mathbf v_6,\mathbf v_7,\mathbf v_8\}$, while the ground truth is known until $t=8$ and the feedback of forecasting performance arrives with 3-step delay.
  (b) A temporal gap always exists between new training samples $\mathcal D_{t-}$ and the test sample $\mathbf X_t$, where concept drift may occur.
  }
  \label{fig:gap}
\end{figure}

As illustrated in Fig.~\ref{fig:gap}, an online learning task inherently has a $H$-step \textit{feedback delay} in time series forecasting, resulting in a temporal gap (at least $H$ steps) between available training samples and the test sample. Formally, at each online time $t$, the current horizon window $\mathbf Y_t$, which has not been observed yet, has an overlap with $\mathbf Y_{t-H+1}, \dots, \mathbf Y_{t-1}$. These samples cannot provide complete forecasting feedback and supervision for the performant, prevailing time series forecasting models that predict $H$-step values directly~\cite{DLinear, PatchTST, iTransformer}. 
Hence, the available training samples for online learning at time $t$ is $\mathcal D_{t-} = \{(\mathbf X_{t'}, \mathbf Y_{t'})\mid t'\le t-H\}$, which has a temporal distance to the test sample $(\mathbf{X}_{t}, \mathbf{Y}_{t})$. 
The temporal gap issue suggests that addressing concept drift by fitting the forecast model to the recent training samples may be insufficient, since concept drift may also occur over the $H$-step temporal gap between the recent training samples and the current test sample.

To investigate the impact of this temporal gap on forecasting performance, we examine two online learning strategies using real-world time series datasets (detailed in Section~\ref{sec:analysis}). The first strategy fine-tunes model parameters using the latest available training sample at time $t$, which is $(\mathbf X_{t-H}, \mathbf Y_{t-H})$. The second strategy~\cite{FSNet, OneNet} omits the feedback delay and adapts the model using $(\mathbf X_{t-1}, \mathbf Y_{t-1})$, which is infeasible in practice due to future information leakage. 
The empirical results show that the average forecast error of the first strategy is approximately double that of the second strategy, indicating that even the latest observed time series pattern differs from the test sample notably. Moreover, the performance gap between the two strategies becomes more significant with a longer horizon, suggesting that the concept drift becomes more pronounced over longer time intervals. These findings highlight the presence of a substantial concept drift between the practical training samples and the test sample, limiting the effectiveness of current online learning techniques that adapt model parameters to potentially outdated concepts and leave the concept drift unresolved. 

In this paper, we aim to answer the question: how can we effectively adapt the forecast model to each test sample and boost online time series forecasting performance against ongoing concept drifts?
A straightforward idea is to estimate a sketch of the latent concept of the test data and customize model parameters for the estimated concept. However, it is challenging to train a model adapter (e.g., a mapping function) that directly generates model parameters based on the estimated concept of the test sample.
First, test data may bear a new concept that is out of the distribution of the historical time series. There is no experience in generating optimal parameters for such an out-of-distribution concept. Second, a parameterized model adapter that maps concepts to model parameters (or their updates) essentially lies in a parameter space of $\mathbb R^{m\times n}$, where $m$ is the concept embedding dimension and $n$ is the number of model parameters. Since advanced time series forecasting models often involve a high quantity of parameters (e.g., 1 million~\cite{PatchTST, iTransformer}), the model adapter is hard to optimize and easily suffers from overfitting due to the huge parameter size and limited test data.

To address these issues, we propose \icon\modelns, a \underline{\textbf{PRO}}a\textbf{\underline{C}}tiv\textbf{\underline{E}} mod\textbf{\underline{E}}l a\underline{\textbf{D}}aptation framework that responds to the concept drift before forecasting the test sample. 
In scenarios where optimal model parameters change with time, we posit that such parameter changes are affected by the drift across a latent concept space, \text{i.e.}, there may exist learnable relationships between the direction and degree of concept drift and those of parameter changes. In light of this, we propose to infer parameter changes based on concept drift, rather than directly generating a whole model.
Specifically, \model begins with a forecast model that has learned new training samples. Given a test sample, \model exploits latent features that encapsulate the concept drift between the training samples and the test sample. In response to the undergoing concept drift, \model generates parameter adjustments via bottleneck layers, which yield a compact set of adaptation coefficients that rescale all model parameters. In this way, our solution eschews direct mapping from concept drift to per-parameter adjustments, instead opting for a more nuanced and efficient strategy. We expect the adapted model with rescaled parameters to approach the optimum for the test sample. To enhance the generalization ability of \modelns, we shuffle historical data to synthesize diverse concept drifts, on which we train \model to learn the relationships between concept drifts and desirable parameter adjustments. When confronted with a reoccurring concept drift that has been learned among the synthetic concept drifts, \model can tailor the model to the test sample. 

The contributions of our work can be summarized as follows.
\begin{itemize}[leftmargin=*]
  \item We highlight that time series forecasting has an inherent feedback delay issue, and we provide an empirical analysis to demonstrate the presence of concept drift between the newly acquired training sample and the test sample. Empirically, such concept drift is more significant with a longer forecast horizon, while it remains unresolved in existing online learning methods. 
  \item We propose \icon\modelns, a proactive model adaptation framework for online time series forecasting under concept drift. \model first estimates the latent representation of concept drift and then customizes parameter adjustments that adapt the parameters to the concept of the test sample in advance of online prediction. 
  \item To improve the generalization ability, we randomly shuffle historical data and synthesize diverse concept drifts to train our framework. In this way, \model has the potential to handle new concepts during the online phase. 
  \item Extensive experiments on five real-world time series datasets demonstrate that \model remarkably reduces the average forecast error of different forecast models by a large margin of 21.9\%. Moreover, \model outperforms existing online model adaptation methods by an average of 10.9\% and keeps high efficiency.
\end{itemize}

\section{Definitions and Problem}~\label{sec:Preliminaries}


Consider $N$ distinct variates ($N \in \mathbb{N}^+$). Let ${\bf v}_t\in \mathbb R^{N}$ denote the $N$-dimensional observation values at time $t$. 
A time series is a sequence of observation vectors in time order, i.e., ${\bf v}_1, {\bf v}_2,\cdots$. 

\begin{definition}[Time Series Forecasting]\label{def:forecast}
Let $L$ be the lookback window size and $H$ be the horizon window size, where $L,H\in \mathbb{N}^+$. At time $t$, a forecast model $\mathcal F$ parameterized by $\boldsymbol\theta$ takes the past observations $\mathbf X_t=[{\bf v}_{t-L+1},\cdots, {\bf v}_{t}]\in \mathbb{R}^{N\times L}$ as input features to predict the future $H$-step values, denoted as $\mathbf Y_t=[{\bf v}_{t+1},\cdots, {\bf v}_{t+H}]\in \mathbb{R}^{N\times H}$.
The training objective is to optimize the model parameters $\boldsymbol\theta$ such that the loss function $\|\hat{\mathbf Y}_t - {\mathbf Y}_t\|_2^2$ is minimized, where $\hat{\mathbf Y}_t\in \mathbb{R}^{N\times H}$ denotes the predicted values in the horizon window. 
\end{definition}

Typically, in multi-step forecasting where $H>1$, there are two primary strategies to generate the predictions ${\bf \hat Y}_t$ at each time $t$.
The first strategy performs \emph{iterative forecasting}. That is, the model forecasts one step ahead and uses its prediction to forecast the next step, repeating this process for the entire horizon. Despite its simplicity, this strategy suffers from significant error accumulation over long horizons~\cite{DLinear}. 
The second strategy adopts \emph{direct forecasting} where the model makes all the predictions ${\bf \hat Y}_t$ simultaneously.
Note that the two strategies require different output modules or layers in the forecast model.
Recent works~\cite{DLinear, PatchTST} have shown that direct forecasting tends to outperform the iterative method, particularly for longer forecast horizons. Hence, this paper adopts the direct forecasting strategy where a sample $\mathbf X_t$ is considered valid for training if all the $H$-step values in $\mathbf Y_t$ are known~\cite{SOLID}. 

In online forecasting scenarios, time series data are observed sequentially. Due to the dynamic nature of real-world processes, underlying data distributions are subject to constant change. Consequently, a forecast model trained on historical data may encounter difficulties when confronted with new, evolving patterns. This issue is referred to as the \emph{concept drift} challenge, which can substantially affect model performance over time~\cite{SurveyConcepDrift}.
To mitigate concept drifts, it is crucial to adapt the forecast model continuously to assimilate new concepts presented in the incoming time series.
%
Formally, the online model adaptation problem is defined as follows.

\begin{definition}[Online Model Adaptation]\label{def:adapt}
At time $t$, online model adaptation activates a model adapter $\mathcal A$ that produces adapted model parameters $\widehat{\boldsymbol\theta}_{t}$ based on available observations $\{\mathbf v_1, \cdots, \mathbf v_t\}$, where $\widehat{\boldsymbol\theta}_{t}$ is expected to be close to the optimal parameters ${\boldsymbol{\theta}}_t^*$. Then, we forecast $\mathbf Y_t$ by $\mathcal F(\mathbf X_t; \widehat{\boldsymbol{\theta}}_t)$.
\end{definition}



At a high level, existing model adaptation methods~\cite{FSNet, OneNet, SOLID} select some observed data $\mathcal{D}_{t-} \subset \{(({\mathbf X}_{t'}, {\mathbf Y}_{t'})\mid t\le t-H\} $ as training samples and utilize the forecasting feedback (\textit{e.g.}, forecasting errors and gradients \textit{w.r.t.} $\mathcal{D}_{t-}$) to update the model parameters. Their adapted parameters tend to align with the patterns or concepts present in ${\mathcal D}_{t-}$.
However, it is crucial to recognize that the concepts present in ${\mathcal D}_{t-}$ may not necessarily reflect that of the test sample $({\mathbf X}_{t}, {\mathbf Y}_{t})$ due the horizon time span $H$. To this end, the model optimized on ${\mathcal D}_{t-}$ might still be susceptible to concept drift, potentially resulting in suboptimal predictions at time $t$ (see details in Section~\ref{sec:analysis}).


In what follows, we provide empirical analysis that illustrates the presence of concept drift between ${\mathcal D}_t$ and the test sample $({\bf X}_t, {\bf Y}_t)$, revealing the limitations of existing model adaptation techniques. 

\section{Empirical Analysis and Motivation}\label{sec:analysis}

In this section, we present an empirical study examining how concept drift between newly acquired training data and test data affects the effectiveness of existing online learning approaches for different time series forecasting models. This analysis highlights the need for developing proactive strategies to adapt forecast models.


\subsection{Datasets}\label{sec:analysis:datasets}
Following prior works~\cite{OneNet, FSNet}, we use five real-world time series datasets, including ETTh2, ETTm1, Weather, ECL, and Traffic. We provide their detailed descriptions in Appendix~\ref{sec:datasets} and statistical information in Table~\ref{tab:dataset}. 
We also adhere to the evaluation settings of FSNet~\cite{FSNet} and OneNet~\cite{OneNet}, where each dataset is chronologically divided into training/validation/test sets by the ratio of 20:5:75. 


\subsection{Baseline Variants}~\label{sec:variants}
We consider the following there time series forecasting models.
\begin{itemize}[leftmargin=*]
    \item FSNet~\cite{FSNet}: FSNet is built upon a TCN~\cite{TCN} backbone and further develops an advanced updating structure that facilitates fast adaptation to concept drift.
    \item OneNet~\cite{OneNet}: OneNet is an ensemble model that dynamically combines two forecasters, one focused on temporal dependence and one focused on channel dependence. We follow its official implementation where the two forecasters are built upon FSNets.
    \item PatchTST~\cite{PatchTST}: PatchTST is one of the state-of-the-art time series forecasters, which models temporal patterns by Transformer.
\end{itemize}
We pre-train the forecast models on the training set and then perform online learning across the validation set and test set. For each of the models, we compare two online learning techniques below.
\begin{itemize}[leftmargin=*]
    \item \textbf{Practical: } At each time $t$, before forecasting $\mathbf Y_t$, we perform predictions on $\mathbf X_{t-H}$ and use the Mean Square Error (MSE) between the ground truth $\mathbf Y_{t-H}$ and the prediction $\hat{\mathbf Y}_{t-H}$ to update the model by one-step gradient descent.
    \item \textbf{Optimal: } At each time $t$, before forecasting $\mathbf Y_t$, we assume the ground truth $\mathbf Y_{t-1}$ is available without any delay. We calculate the previous forecast loss on the last sample $\mathbf X_{t-1}$ (\textit{i.e.}, MSE between $\mathbf Y_{t-1}$ and $\hat{\mathbf Y}_{t-1}$) to update the model by one-step gradient descent.
\end{itemize}
It is important to notice that the \textit{Optimal} strategy uses the most relevant and recent information for the prediction on the test sample ${\bf X}_t$. However, it is infeasible in practice as it requires knowledge of future data. The performance gap between this ideal strategy and the Practical strategy reveals the presence of concept drift that is not adequately addressed by existing online learning techniques. 
To maintain a fair comparison, both strategies fine-tune all the model parameters using one training sample each time.


\subsection{Evaluation Metrics}
Following previous works~\cite{FSNet, Informer, DLinear}, we evaluate the forecasting performance by two commonly used metrics as defined below:
$$
\text{Mean Square Error (MSE)}=\frac{1}{N|T_\text{test}|}\sum_{t\in T_\text{test}}\|\hat{\mathbf Y}_t - {\mathbf Y}_t\|_2^2, 
$$
$$
\text{Mean Absolute Error (MAE)}=\frac{1}{N|T_\text{test}|}\sum_{t\in T_\text{test}}\|\hat{\mathbf Y}_t - {\mathbf Y}_t\|_1,
$$
where $T_\text{test}$ represents all time steps of the test set and $\hat{\mathbf Y}_t \in \mathbb R^{N\times H}$ denotes the predicted result at time $t$. A lower error indicates higher forecasting performance. Each experiment is repeated 3 times with different random seeds and we report the average results. Due to the page limit, we provide the MAE results in Appendix~\ref{sec:gap-mae}.

Additionally, we quantify the performance gap between the \textit{Practical} and the \textit{Optimal} by the following equation:
$$
\Delta_{\text{MSE}}=\frac{\text{MSE}_{Practical} - \text{MSE}_{Optimal}}{\text{MSE}_{Optimal}} \times 100\%.
$$

\begin{table*}[t]
	\centering
	\caption{MSE results of two online learning strategies for time series forecasting. We report the MAE results in Appendix~\ref{sec:gap-mae}.  }\label{tab:mse gap}
  \resizebox{0.95\linewidth}{!}{
\begin{tabular}{cc|ccc|ccc|ccc|ccc|ccc}
\toprule
 & \multirow{2}{*}{\diagbox[width=2cm,height=0.85cm,innerrightsep=2pt,innerwidth=1.5cm]{}{\small \textbf{Dataset}\\$H$}} & \multicolumn{3}{c|}{\textbf{ETTh2}} & \multicolumn{3}{c|}{\textbf{ETTm1}} & \multicolumn{3}{c|}{\textbf{Weather}} & \multicolumn{3}{c|}{\textbf{ECL}} & \multicolumn{3}{c}{\textbf{Traffic}} \\
\multicolumn{1}{c}{\textbf{Model}} &  & \textbf{24} & \textbf{48} & \textbf{96} & \textbf{24} & \textbf{48} & \textbf{96} & \textbf{24} & \textbf{48} & \textbf{96} & \textbf{24} & \textbf{48} & \textbf{96} & \textbf{24} & \textbf{48} & \textbf{96} \\ \midrule
\multirow{3}{*}{FSNet} & Optimal & 0.687 & 0.846 & \multicolumn{1}{l|}{1.087} & 0.584 & 0.724 & 0.706 & 0.786 & 1.039 & 1.107 & 5.668 & 5.811 & 5.979 & 0.362 & 0.464 & 0.501 \\
 & Practical & 3.079 & 4.247 & 6.213 & 0.763 & 0.923 & 1.003 & 0.875 & 1.338 & 1.755 & 6.143 & 9.435 & 12.941 & 0.458 & 0.517 & 0.574 \\
 & $\Delta_{\text{MSE}}$ & 348\% & 402\% & 472\% & 31\% & 27\% & 42\% & 11\% & 29\% & 59\% & 8\% & 62\% & 116\% & 26\% & 11\% & 15\% \\ \midrule
\multirow{3}{*}{OneNet} & Optimal & 0.532 & 0.609 & \multicolumn{1}{l|}{0.993} & 0.173 & 0.173 & 0.189 & 0.379 & 0.397 & 0.415 & 2.074 & 2.201 & 3.089 & 0.364 & 0.400 & 0.438 \\
 & Practical & 2.965 & 4.892 & 8.257 & 1.335 & 1.547 & 1.040 & 0.861 & 1.182 & 1.484 & 9.757 & 10.847 & 15.932 & 0.462 & 0.544 & 0.591 \\
 & $\Delta_{\text{MSE}}$ & 457\% & 703\% & 732\% & 672\% & 794\% & 452\% & 128\% & 198\% & 257\% & 370\% & 393\% & 416\% & 27\% & 36\% & 35\% \\ \midrule
\multirow{3}{*}{PatchTST} & Optimal & 1.664 & 2.596 & 4.367 & 0.245 & 0.240 & 0.236 & 0.546 & 0.565 & 0.547 & 4.085 & 4.699 & 5.602 & 0.358 & 0.341 & 0.348 \\
 & Practice & 2.092 & 3.434 & 5.770 & 0.455 & 0.598 & 0.674 & 0.735 & 0.979 & 1.261 & 4.143 & 4.762 & 5.791 & 0.376 & 0.380 & 0.398 \\
 & $\Delta_{\text{MSE}}$ & 26\% & 32\% & 32\% & 86\% & 149\% & 185\% & 35\% & 73\% & 130\% & 1\% & 1\% & 3\% & 5\% & 11\% & 15\% \\ 
 \bottomrule
  \end{tabular}
  }
  \begin{tablenotes}
  \footnotesize
      \item \textsuperscript{\textdaggerdbl} We augment FSNet and OneNet with RevIN~\cite{RevIN} for better performance. The results differ from those in the original paper. We explain the reasons in Appendix~\ref{sec:setting}.
\end{tablenotes}
\end{table*}

\subsection{Key Observations}
Table~\ref{tab:mse gap} shows the MSE results of two strategies. 
We have the following major observations. 
\textbf{First}, the \textit{Practical} strategy performs worse than \textit{Optimal} by an average of 107\% on different models. This significant difference suggests that the most recently observed pattern in $(\mathbf X_{t-H}, \mathbf Y_{t-H})$ substantially differs from $(\mathbf X_{t-1}, \mathbf Y_{t-1})$ and fails to reflect the concept of the test sample $(\mathbf X_{t}, \mathbf Y_{t})$. In other words, the adapted forecast models are still vulnerable to the concept drift caused by the feedback delay issue. 
\textbf{Second}, we can observe the general tendency for the performance gap to become more significant as the forecast horizon increases. This could be attributed to the increasing temporal distance between $\mathbf X_{t-H}$ and $\mathbf X_{t}$.
\textbf{Third}, OneNet demonstrates the best performance when utilizing the \textit{Optimal} strategy. However, its effectiveness drastically diminishes in all the cases when employing the \textit{Practical} strategy, even becoming the worst on the ETTm1 and Traffic datasets. 
The reason is that OneNet is an ensemble model with a large number of parameters, which increases the overfitting risk on the new training samples. 
\textbf{Fourth}, PatchTST, recognized as the leading time series forecasting model, reports much smaller $\Delta_{\text{MSE}}$ than the other models. 
When using the \textit{Practical} strategy, PatchTST outperforms all the other models. However, it still falls short of the performance achieved by OneNet under the \textit{Optimal} strategy. To sum up, there is still considerable room for performance improvement if we can address the possible concept drift between each test sample and the latest available training data during online learning.   

\section{The Proposed Framework: \model}\label{sec:method}
In this section, we first discuss our idea and provide an overview of the proposed \model solution. We then elaborate on two major steps, namely concept drift estimation and proactive model adaptation. Finally, we describe the end-to-end training scheme that improves the model's adaptation ability against concept drifts.

\subsection{Solution Overview}\label{sec:overview}
The ultimate goal of \model is to close the gap between the newly acquired training data and the test sample and boost performance against concept drift caused by the feedback delay issue. To achieve this, a simple solution is to extract the latent concepts from all historical samples and learn a mapping function between each concept and optimal model parameters \textit{w.r.t.} the historical sample. As such, one can extract concept from each test sample and use the mapping function to directly generate possibly optimal parameters. However, the online phase may include new concepts that are out of the historical data distribution. The simple solution can fail in this case, since it has not learned the relationship between out-of-distribution (OOD) concepts and corresponding optimal parameters. 

To address the problem, we propose to map \emph{concept drifts} to \emph{parameter shifts}. We assume that the direction and degree of the drift over the concept space can reflect a possible direction and magnitude of parameter shifts over the parameter space, informing how to make an adaptation. Specifically, given a model that fits new training samples, \model exploits latent features from the training samples and the current test sample, estimates the undergoing concept drift, and accordingly predicts parameter shifts. 

\begin{figure}[t]
  \centering
  \includegraphics[width=.85\linewidth]{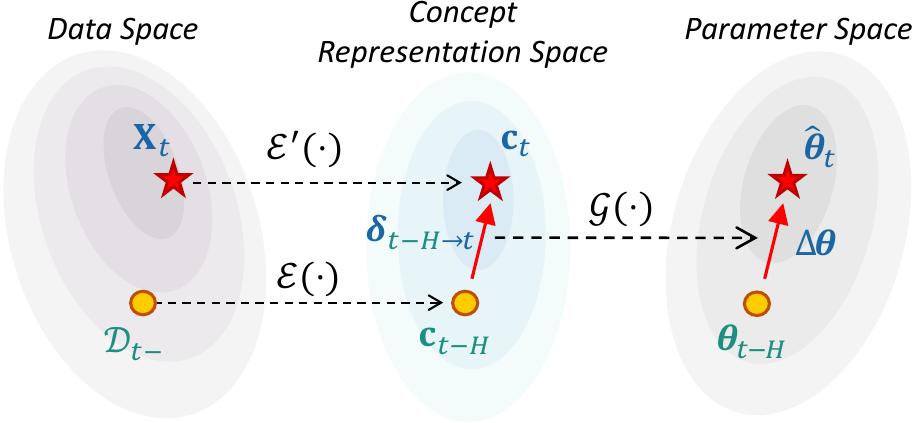}
  \caption{Illustration of the mapping from time series data change to parameter shift. Concept encoders $\mathcal E$ and $\mathcal E'$ encode time series into concept representations. Adaptation generator $\mathcal G$ decodes the conceptual difference into parameter shift.}
  \label{fig:shift}
\end{figure}

Following existing online model adaptation methods~\cite{FSNet, OneNet}, we use one training sample for model updating at each time $t$, \textit{i.e.}, $\mathcal D_{t-}={(\mathbf X_{t-H}, \mathbf Y_{t-H})}$. With the model updated on $\mathcal D_{t-}$, we estimate the concept drift between $\mathcal D_{t-}$ and $\mathbf X_t$, predict potential shifts in the parameter space, and accordingly make parameter adjustments.
Formally, our \model solution consists of four key steps at each time $t$ as listed below.
\begin{enumerate}[leftmargin=*]
    \item \textbf{Online Fine-tuning.\label{step:fine-tuning}} Given a forecast model $\mathcal F$ parameterized by $\boldsymbol{\theta}_{t-H-1}$, we redo forecasting by $\hat{\mathbf{Y}}_{t-H} = \mathcal F(\mathbf{X}_{t-H}; \boldsymbol{\theta}_{t-H-1})$. Next, we use the forecast error $\|\hat{\mathbf{Y}}_{t-H} - {\mathbf{Y}}_{t-H}\|^2_2$ to update $\boldsymbol{\theta}_{t-H-1}$ into $\boldsymbol{\theta}_{t-H}$ by gradient descent. The subscript of $\boldsymbol\theta$ indicates that the parameters have fit $(\mathbf X_{t-H}, \mathbf Y_{t-H})$.
    \item \textbf{Concept drift estimation.} Given $\mathcal{D}_{t-}$ and the test sample $\mathbf{X}_{t}$, we feed them into two concept encoders $\mathcal E$ and $\mathcal E'$ that extract concept representations denoted as $\mathbf{c}_{t-H}\in \mathbb R^{d_c}$ and $\mathbf{c}_{t}\in \mathbb R^{d_c}$, respectively. Then, we estimate the hidden state of the concept drift between $\mathbf{X}_{t-H}$ and $\mathbf{X}_{t}$ by $\boldsymbol{\delta}_{t-H\to t}$, where $\boldsymbol{\delta}_{t-H\to t} = \mathbf{c}_{t} - \mathbf{c}_{t-H}$.
    \item \textbf{Proactive model adaptation.} Given the estimated concept drift $\boldsymbol{\delta}_{t-H\to t}$ and the parameters $\boldsymbol{\theta}_{t-H}$, we employ an adaptation generator $\mathcal G$ to generate parameter shifts $\Delta\boldsymbol{\theta}$, adjusting $\boldsymbol{\theta}_{t-H}$ into $\hat{\boldsymbol{\theta}}_{t}$ as illustrated in Fig.~\ref{fig:shift}. 
    \item \textbf{Online forecasting.} Finally, the adapted model yields predictions by $\hat{\mathbf{Y}}_{t} = \mathcal F(\mathbf{X}_{t}; \hat{\boldsymbol{\theta}}_{t})$. At the next time $t+1$, the parameters will be reset to $\boldsymbol{\theta}_{t-H}$.
\end{enumerate}

\begin{figure}[t]
  \centering
  \setlength{\belowcaptionskip}{-5pt}
  \subcaptionbox{\label{fig:concept}Concepts}
  {
    \includegraphics[width=.5\linewidth]{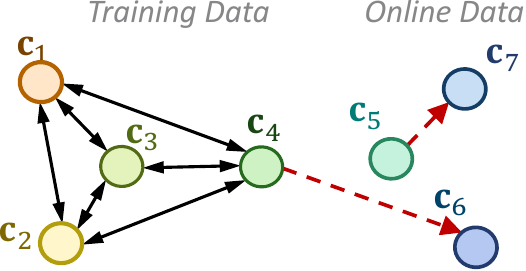}
  }\hspace{1.5pt}
  \subcaptionbox{\label{fig:drift}Vectors of concept drifts}
  {
    \includegraphics[width=.44\linewidth]{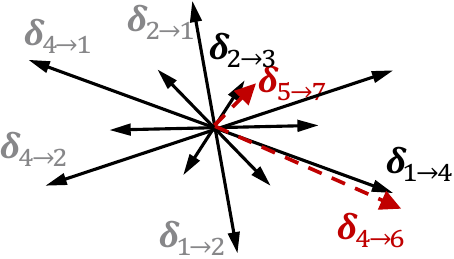}
  }
  \caption{Example of synthetic concept drifts in historical training data and online concept drifts when $H=2$. The circles represent concepts, while the arrows represent concept drifts.
  }
  \label{fig:vs}
\end{figure}

The rationale of our solution is that we can synthesize diverse concept drifts based on historical data to train our model adapter. 
Fig.~\ref{fig:vs} shows an example of the historical training data including four samples with their concepts denoted by $\mathbf c_1, \cdots, \mathbf c_4$ respectively. Our idea is to shuffle the order of samples and train our model adapter on the synthetic concept drifts between each pair of samples. Though the concepts of the future test samples (\textit{i.e.}, $\mathbf c_5, \cdots, \mathbf c_7$) are out-of-distribution, the concept drift patterns $\boldsymbol \delta_{4\to 6}$ and $\boldsymbol \delta_{5\to 7}$ are similar to $\boldsymbol \delta_{1\to 4}, \boldsymbol \delta_{2\to 3}$ which have been learned in training data. Given such recurring concept drifts, we can generate appropriate parameter shifts by experience.

\subsection{Concept Drift Estimation}

In the literature, there are numerous concept drift detection methods~\cite{SurveyConcepDrift} that estimate the degree of concept drift to decide when to adapt the model. The degree can be estimated by the changes in forecast error~\cite{OSAKA}, distribution distance~\cite{AdaRNN}, prediction uncertainty~\cite{METER}, \textit{etc}. Nevertheless, the degree as a scalar in $\mathbb R$ is not informative and cannot indicate how to adapt the model. Therefore, we propose to model a high-dimensional representation vector in $\mathbb R^{d_c}$ that characterizes both the degree and the direction of the concept drift, where $d_c$ is the representation dimension.

First, we devise a concept encoder $\mathcal E$ that extracts concept representation $\mathbf c_{t-H}$ from the latest training sample $\mathcal D_{t-}={(\mathbf X_{t-H}, \mathbf Y_{t-H})}$. Let $\mathbf X_{t-H}^{(i)} \in \mathbb R^{L}$ denote the time series of the $i$-th variate. Given all observations of $N$ variates at time $t$, we encode $\mathbf c_{t-H}$ by the following equation:
\begin{equation}\label{eq:avg}
    \mathbf{c}_{t-H} = \mathcal{E}(\mathcal D_{t-}) = \mathtt{Average}\left(\{\mathtt{MLP}(\mathbf X_{t-H}^{(i)} \| \mathbf Y_{t-H}^{(i)})\}_{i=1}^N\right) \in \mathbb R^{d_c},
\end{equation}
where $\mathtt{MLP}$ extracts $d_c$ latent features (\textit{e.g.}, mean and standard deviation) from each univariate time series, and $\mathtt{Average}$ yields the average latent features as a global concept that affects all variates.

Likewise, we employ another MLP to extract latent features from $\mathbf X_{t}$, and the encoder $\mathcal E'$ extracts $\mathbf c_{t}$ by
\begin{equation}
    \mathbf{c}_t = \mathcal{E'}(\mathbf X_{t}) = \mathtt{Average}\left(\{\mathtt{MLP'}(\mathbf X_{t}^{(i)})\}_{i=1}^N\right) \in \mathbb R^{d_c}.
\end{equation}
It is noteworthy that $\mathcal E'$ has the potential to estimate the hidden state of $\mathbf Y_{t}$. When $L \ge kH$ and $k\ge 2$, $\mathbf{X}_t$ itself contains a sequence of horizon windows, \textit{i.e.}, $\{\mathbf Y_{t-kH}, \cdots, \mathbf Y_{t-H}\}$. In this case, $\mathtt{MLP'}$ can learn the hidden state of each observed horizon window, exploit the temporal evolution pattern across horizon windows, and extrapolate the hidden state of the next horizon window $\mathbf Y_{t}$. 

Next, we estimate the concept drift between time $t-H$ and time $t$ by the concept difference, \textit{i.e.}, $\boldsymbol{\delta}_{t-H\to t} = \mathbf{c}_{t} - \mathbf{c}_{t-H} \in \mathbb R^{d_c}$. 


\subsection{Proactive Model Adaptation}

As illustrated in Fig.~\ref{fig:shift}, we assume the concept representation space and the parameter space to have some learnable relationships, where the estimated concept drift $\boldsymbol{\delta}_{t-H\to t}$ can indicate the direction that the parameters $\boldsymbol\theta_{t-H}$ should shift to. 

\begin{figure}[t]
  \centering
  \includegraphics[width=0.9\linewidth]{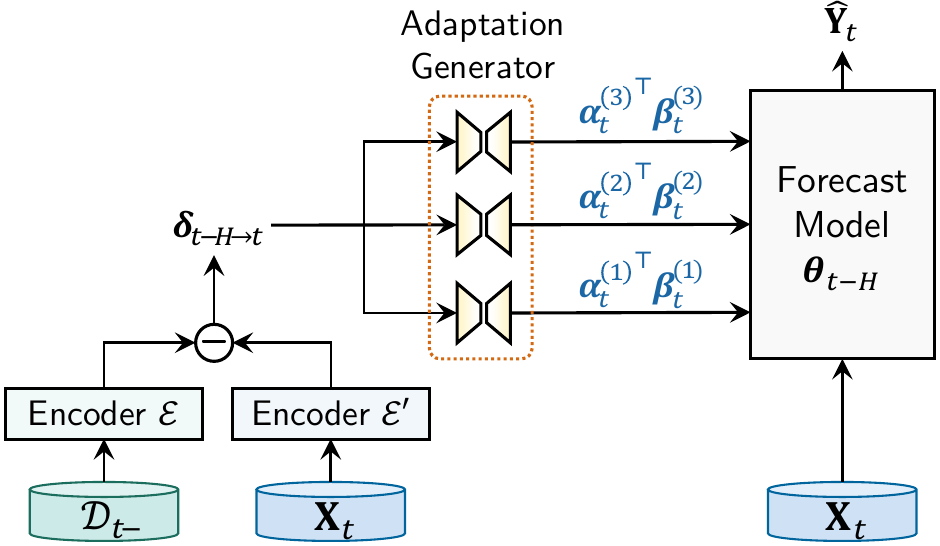}
  \caption{Overview of the model adapter in \model that comprises two concept encoders and an adaptation generator.}
  \label{fig:framework}
\end{figure}

Technically, it is non-trivial to decode the concept drift representation into an appropriate parameter shift.
As the parameter space is often of a huge dimension, it is tough to search for an optimal mapping function between the concept space and the parameter space. Also, a simple mapping function may require too many additional parameters, leading to costly memory overhead. For instance, when $d_c=100$ and the number of model parameters is 1 million, a trivial method is to learn a fully connected layer of 100 million parameters that maps $\boldsymbol{\delta}_{t-H\to t}$ to parameter shifts $\Delta\boldsymbol{\theta}$. 
To address these problems, as depicted in Fig.~\ref{fig:framework}, we devise an adaptation generator $\mathcal G$ that employs bottleneck layers to produce a small number of adaptation coefficients $\boldsymbol{\alpha}^{(\ell)}$ and $\boldsymbol{\beta}^{(\ell)}$ for each the $\ell$-th layer.

For simplicity, we take a linear transformation matrix as an example and denote the $\ell$-th layer as $\boldsymbol\theta^{(\ell)}_{t-H}\in \mathbb R^{d_{\textit{in}}\times d_{\textit{out}}}$. Given a small hyperparameter $r$ ($r\ll \min(d_{\textit{in}}, d_{\textit{out}})$), we compute adaptation coefficients by
\begin{equation}    
[{\boldsymbol{\alpha}}_{t}^{(\ell)}, {\boldsymbol{\beta}}_{t}^{(\ell)}]= {\mathbf{W}_2^{(\ell)}}^{\top}\left(\sigma\left(\mathbf{W}_1^{(\ell)^{\top}} \boldsymbol{\delta}_{t-H \to t} + \mathbf b^{(\ell)} \right)\right)+\mathbf 1,
\end{equation}
where $\mathbf{W}_1^{(\ell)} \in \mathbb R^{r\times d_c}$, $\mathbf b^{(\ell)} \in \mathbb R^{r}$, $\sigma$ is the sigmoid activation, and $\mathbf{W}_2^{(\ell)} \in \mathbb R^{(d_{\textit{in}} + d_{\textit{out}})\times r}$. ${\boldsymbol{\alpha}}_{t}^{(\ell)}\in \mathbb R^{d_\textit{in}}$ and ${\boldsymbol{\beta}}_{t}^{(\ell)} \in \mathbb R^{d_\textit{out}}$ are initialized as $\mathbf 1$ by initializing $\mathbf{W}_2^{(\ell)}$ as zeros. 

To further reduce parameters, we share $\mathbf{W}_1^{(\ell)}$ and $\mathbf{W}_2^{(\ell)}$ across layers of the same type (\textit{e.g.}, up projection layers in all Transformer blocks), while we learn an individual bias term $\mathbf b^{(\ell)}$ to customize distinct coefficients for each layer. Compared with a fully connected layer, we reduce the total parameters of the adaptation generator from $\mathcal{O}(\mathcal L d_c d_{\textit{in}}d_{\textit{out}})$ to $\mathcal{O}(r(\mathcal L + d_c + d_{\textit{in}} + d_{\textit{out}}))$, where $\mathcal L$ is the number of model layers.

Finally, we derive the adapted parameters used in online forecasting at time $t$ by:
\begin{equation}\label{eq:final adapt}
        \widehat{\boldsymbol{\theta}}_{t}^{(\ell)}
    = \left( {{\boldsymbol\alpha}_{t}^{(\ell)}}^\top{\boldsymbol{\beta}}_{t}^{(\ell)}\right)  \odot\boldsymbol{\theta}_{t-H}^{(\ell)},
\end{equation}
where $\odot$ is the element-wise product. In other words, we generate model adjustments $\Delta\boldsymbol{\theta}^{(\ell)}= \left( {{\boldsymbol\alpha}_{t}^{(\ell)}}^\top{\boldsymbol{\beta}}_{t}^{(\ell)}-\mathbf 1\right)  \odot\boldsymbol{\theta}_{t-H}^{(\ell)}$. 

We also apply our method to convolution filters and bias terms. For a convolution filter $\boldsymbol{\theta}^{(\ell)} \in \mathbb R^{d_{\textit{in}}\times d_{\textit{out}}\times d_{\kappa}}$ with $d_{\kappa}$ kernels, we multiply each kernel with ${\boldsymbol{\alpha}_{t}^{(\ell)}}^\top\boldsymbol{\beta}_{t}^{(\ell)}$. For a bias term $\boldsymbol{\theta}^{(\ell)} \in \mathbb R^{d_{\textit{out}}}$, we only generate $\boldsymbol{\beta}_{t}^{(\ell)}$.

\subsection{Mini-batch Training}~\label{sec:training}

Given abundant historical data, we shuffle them to synthesize diverse concept drifts and train our model adapter on them. 
To improve the training efficiency, we randomly select multiple samples as a mini-batch, adapt the forecaster towards each sample concurrently, and compute the average forecast loss.

Let $\mathcal B_{k} = \{\mathbf X_{k,j}, \mathbf Y_{k,j}\}_{j=kB+1}^{kB+B}$ represent the $k$-th mini-batch of $B$ samples collected from different time. For the last mini-batch $\mathcal B_{k-1}$, our concept encoder $\mathcal E'$ extracts latent features of all samples and their average  $\mathbf c_{k-1}$ is considered as the concept representation of $\mathcal B_{k-1}$. For each $\mathbf X_{k,j}$ in $\mathcal B_{k}$, we estimate the concept drift from $\mathcal B_{k-1}$ to $\mathbf X_{k,j}$ individually and generate the corresponding adaptation coefficients $\boldsymbol{\alpha}_{k,j}^{(\ell)}, \boldsymbol{\beta}_{k,j}^{(\ell)}$. Note that we do not iteratively make different versions of adapted models (\textit{i.e.}, $\{\widehat{\boldsymbol{\theta}}_{k,j}\}_{j=kB+1}^{kB+B}$) which consumes more GPU memory and more time cost. Instead, we only preserve one model obtained from the last mini-batch (denoted as $\boldsymbol{\theta}_{k-1}$) and simultaneously handle multiple data batches.

For brevity, we omit the superscript $(\ell)$ in the following equations. Let $\mathbf{h}_{k,j}\in \mathbb R^{d_{\textit{in}}}$ denote the intermediate inputs of the $\ell$-th layer. Given adaptation coefficients $\boldsymbol{\alpha}_{k,j},\boldsymbol{\beta}_{k,j}$, the linear transformation via adapted parameter $\widehat{\boldsymbol{\theta}}_{k,j}$ is reformulated as
\begin{equation}
\begin{aligned}\label{eq:vectorization}
    \widehat{\boldsymbol{\theta}}_{k,j}^{\top} \mathbf{h}_{k,j}
    &= \left(({\boldsymbol{\alpha}_{k,j}}^\top\boldsymbol{\beta}_{k,j}) \odot \boldsymbol{\theta}_{k-1}\right)^{\top} \mathbf{h}_{k,j} \\
    & =\boldsymbol{\beta}_{k,j} \odot\left(\boldsymbol{\theta}_0^{\top} ({\boldsymbol{\alpha}_{k,j}} \odot \mathbf{h}_{k,j})\right).
\end{aligned}
\end{equation}
This equation also stands for other parameters such as convolution filters and bias terms. As such, we can directly compute $\boldsymbol{\theta}_0^{(\ell)\top} \mathbf{H}_{k}^\prime$ as usual, where $\mathbf{H}_{k}^\prime = \{{\boldsymbol{\alpha}_{k,j}}\odot\mathbf{h}_{k,j}\}_{j=kB+1}^{kB+B}$. Correspondingly, we multiply the layer outputs with $\boldsymbol{\beta}_{k,j}$ for each sample in parallel.

As such, we obtain the predictions of all samples in $\mathcal B_{k}$, denoted as $\{\hat{\mathbf{Y}}_{k,j}\}_{j=kB+1}^{kB+B}$. Next, we compute the average forecast loss on $\frac{1}{B}\sum_{j=kB+1}^{kB+B} \|\hat{\mathbf{Y}}_{k,j}-\mathbf{Y}_{k,j}\|_2^2$ and use the gradients to update $\boldsymbol{\theta}_{k-1}$ into $\boldsymbol{\theta}_{k}$ and train other additional parameters, including the concept encoders and the adaptation generator. The updated model and the updated adapters will be used in the next mini-batch. During the online phase, we only fine-tune the forecast model and keep the parameters of the model adapter frozen because the ground truth of the test sample is not available.

As our model adapter learns the transition probability $P(\boldsymbol{\theta}_{t} \mid \boldsymbol{\theta}_{s}, \mathbf{X}_{s}, \mathbf{X}_{t})$, we can follow the proof in the previous work~\cite{DRAIN} to show that it is feasible for $\widehat{\boldsymbol{\theta}}_{t}$ in Eq.~(\ref{eq:final adapt}) to yield lower forecasting error than $\boldsymbol{\theta}_{t-H}$ (see Appendix~\ref{sec:theorem}).

\begin{table*}[th!]
    \centering
    \caption{Difference between online model adaptation methods.}
  \resizebox{0.9\textwidth}{!}{  
\begin{threeparttable}
    \begin{tabular}{c|c|c|c}
    \toprule
        \textbf{Method} & \textbf{Training sample(s)} $\mathcal D_{t-}$ \textbf{at time} $t$ &  \makecell[c]{\textbf{Feedback-based model adaptation} \\(How to adapt to $\mathcal D_{t-}$)} & \makecell[c]{\textbf{Proactive model adaptation} \\(How to adapt to $\mathbf X_t$)} \\\midrule
        FSNet~\cite{FSNet} & $(\mathbf X _{t-H},\mathbf Y _{t-H})^{\dagger}$ & \makecell[c]{GD\textsuperscript{\textdaggerdbl} + Adjusting convolutional filters \\based on $\nabla \text{MSE}(\hat{\mathbf Y}_{t-H},\mathbf Y _{t-H})$} & \XSolidBrush \\\midrule
        OneNet~\cite{OneNet} & $(\mathbf X _{t-H},\mathbf Y _{t-H})^{\dagger}$ & \makecell[c]{GD\textsuperscript{\textdaggerdbl} + Adjusting ensemble weights \\based on $\text{MSE}(\hat{\mathbf Y}_{t-H},\mathbf Y _{t-H})$} & \XSolidBrush \\\midrule
        SOLID~\cite{SOLID} & $\left\{(\mathbf X _s,\mathbf Y _s):s\le t-H\wedge \|\mathbf X _s -\mathbf X _t\|_2\in TopK\right\}$ & GD\textsuperscript{\textdaggerdbl} & \XSolidBrush \\\midrule
        \icon\model & $(\mathbf X _{t-H},\mathbf Y _{t-H})$ & GD\textsuperscript{\textdaggerdbl} & \makecell[c]{Adjusting all parameters \\based on $\mathbf X _{t-H}$ and $\mathbf X _t$} \\\bottomrule
    \end{tabular}
    \begin{tablenotes}
    \footnotesize
      \item \textsuperscript{\textdagger} FSNet and OneNet used $(\mathbf X_{t-1}, \mathbf Y_{t-1})$ in their official implementations, while the practical strategy is to use $(\mathbf X_{t-H}, \mathbf Y_{t-H})$ without information leakage.
      \item \textsuperscript{\textdaggerdbl} "GD" refers to applying gradient descent algorithm on $\mathcal D_{t-}$. 
    \end{tablenotes}
    \end{threeparttable}
    }
    \label{tab:method comparison}
\end{table*}

\subsection{Discussion} \label{sec:discussion}

\subsubsection{Comparison with Existing Methods.} 

Prior approaches to online time series forecasting perform model adaptation based on feedback in forecasting previous samples, which mainly focus on learning recent data patterns. In Table~\ref{tab:method comparison}, we compare existing methods and \model in terms of the training data and the model adaptation techniques at each online time $t$. Most methods use one newly acquired training sample, while SOLID~\cite{SOLID} selects more recent samples that share similar lookback windows with the test sample and are assumed to share a similar concept. SOLID and \model simply fine-tune the forecast model by gradient descent, while FSNet and OneNet further generate additional parameter adjustments based on the forecasting feedback. Since the feedback is delayed by $H$ steps and cannot reveal the test concept, we propose a novel step called proactive model adaptation which aims to mitigate the effects of concept drift between the training sample and the test sample. As this core step is orthogonal to existing methods, \model can incorporate the data augmentation technique of SOLID and the feedback-based model adaptation techniques of FSNet and OneNet.

\subsubsection{Time Complexity Analysis.} 
Though the lookback window size $L$ and the number of variates $N$ could be large, our concept drift estimation has a linear complexity w.r.t. them, \textit{i.e.}, $\mathcal O(NLd_c)$ with a relatively small hyperparameter $d_c$.
As for proactive model adaptation, the time complexity is $\mathcal O(r(d_c + d_{\textit{in}}+d_{\textit{out}}) + d_{\textit{in}}d_{\textit{out}})$, which is agnostic to the number of variates. Note we set a rather small bottleneck dimension $r$ (\textit{e.g.}, 32). Hence, our framework is friendly to large-scale multivariate time series with a large $N$. Throughout the online phase, we first adapt the model parameters by Eq.~(\ref{eq:final adapt}), and online forecasting is performed with no additional cost.

\section{Experiments}

\begin{table*}[t]
	\centering
	\caption{MSE of different online methods with horizons in \{24, 48, 96\}. The best results are marked in \textbf{bold}.
 }\label{tab:main_result_mse}
  \resizebox{0.9\textwidth}{!}{
  \begin{tabular}{cc|ccc|ccc|ccc|ccc|ccc}
\toprule
 & \multirow{2}{*}{\diagbox[width=2cm,height=0.85cm,innerrightsep=2pt,innerwidth=1.5cm]{\textbf{\small Method}}{\small \textbf{Dataset}\\$H$}} & \multicolumn{3}{c|}{\textbf{ETTh2}} & \multicolumn{3}{c|}{\textbf{ETTm1}} & \multicolumn{3}{c|}{\textbf{Weather}} & \multicolumn{3}{c|}{\textbf{ECL}} & \multicolumn{3}{c}{\textbf{Traffic}} \\
\multicolumn{1}{c}{\textbf{Model}} &  & \textbf{24} & \textbf{48} & \textbf{96} & \textbf{24} & \textbf{48} & \textbf{96} & \textbf{24} & \textbf{48} & \textbf{96} & \textbf{24} & \textbf{48} & \textbf{96} & \textbf{24} & \textbf{48} & \textbf{96} \\ \midrule
\multirow{6}{*}{TCN} & \textbackslash{} & 3.156 & 4.141 & 6.019 & 1.770 & 1.908 & 1.673 & 3.301 & 9.718 & 8.588 & 57.782 & 57.948 & 58.729 & 0.569 & 0.637 & 0.676 \\
 & GD & 3.223 & 4.408 & 6.291 & 1.015 & 1.172 & 1.142 & 0.868 & 1.346 & 1.799 & 6.315 & 8.176 & 11.385 & 0.452 & 0.509 & 0.576 \\
 & FSNet & 3.079 & 4.247 & 6.213 & 0.763 & 0.923 & 1.003 & 0.875 & 1.338 & 1.755 & 6.143 & 9.435 & 12.941 & 0.458 & 0.517 & 0.574 \\
 & OneNet & 2.965 & 4.892 & 8.257 & 1.335 & 1.547 & 1.040 & 0.861 & 1.182 & 1.484 & 9.757 & 10.847 & 15.932 & 0.462 & 0.544 & 0.591 \\
 & SOLID++ & 3.231 & 4.111 & 6.191 & 0.674 & 0.795 & 0.868 & 0.867 & 1.311 & 1.670 & 5.991 & 7.340 & 9.412 & 0.418 & 0.479 & 0.529 \\
 & \modelns & \textbf{2.908} & \textbf{4.056} & \textbf{5.891} & \textbf{0.531} & \textbf{0.704} & \textbf{0.780} & \textbf{0.707} & \textbf{0.959} & \textbf{1.314} & \textbf{5.907} & \textbf{7.192} & \textbf{9.183} & \textbf{0.413} & \textbf{0.454} & \textbf{0.511} \\ \midrule
\multirow{5}{*}{PatchTST} & \textbackslash{} & 2.880 & 4.103 & 6.178 & 0.549 & 0.726 & 0.769 & 0.746 & 1.006 & 1.311 & 4.207 & 4.942 & 6.042 & 0.378 & 0.386 & 0.403 \\
 & GD & 2.092 & 3.434 & 5.770 & 0.455 & 0.598 & 0.674 & 0.735 & 0.979 & 1.261 & 4.143 & 4.762 & 5.791 & 0.376 & 0.380 & 0.398 \\
 & OneNet & 2.717 & 4.095 & 6.044 & 0.681 & 0.721 & 0.788 & 0.824 & 1.052 & 1.319 & 4.111 & 4.752 & 5.767 & 0.360 & 0.372 & 0.387 \\
 & SOLID++ & 2.648 & 3.925 & 6.148 & 0.447 & 0.580 & \textbf{0.659} & 0.735 & 0.981 & 1.262 & 4.156 & 4.780 & 5.835 & 0.376 & 0.378 & 0.397 \\
 & \modelns & \textbf{1.735} & \textbf{3.114} & \textbf{5.555} & \textbf{0.424} & \textbf{0.577} & 0.660 & \textbf{0.724} & \textbf{0.973} & \textbf{1.261} & \textbf{3.958} & \textbf{4.604} & \textbf{5.635} & \textbf{0.335} & \textbf{0.357} & \textbf{0.376} \\ \midrule
\multirow{5}{*}{iTransformer} & \textbackslash{} & 3.605 & 4.992 & 7.114 & 0.603 & 0.807 & 0.832 & 0.942 & 1.211 & 1.461 & 4.082 & 4.874 & 6.054 & 0.354 & 0.372 & 0.393 \\
 & GD & 2.637 & 4.148 & 6.734 & 0.468 & 0.606 & 0.695 & 0.869 & 1.143 & 1.409 & 4.055 & 4.851 & 5.989 & 0.348 & 0.369 & 0.388 \\
 & OneNet & 2.985 & 4.015 & 6.451 & 0.609 & 0.746 & 0.766 & 0.803 & 1.089 & 1.418 & 4.077 & 4.906 & 6.100 & 0.349 & 0.371 & 0.395 \\
 & SOLID++ & 2.804 & 4.278 & 6.582 & 0.455 & 0.601 & 0.688 & 0.875 & 1.139 & 1.403 & 4.070 & 4.816 & 6.024 & 0.342 & 0.365 & 0.384 \\
 & \modelns & \textbf{2.387} & \textbf{3.969} & \textbf{6.291} & \textbf{0.426} & \textbf{0.561} & \textbf{0.642} & \textbf{0.742} & \textbf{1.015} & \textbf{1.294} & \textbf{3.899} & \textbf{4.647} & \textbf{5.710} & \textbf{0.330} & \textbf{0.353} & \textbf{0.373} \\ \bottomrule
\end{tabular}
}
\end{table*}

\begin{table*}[t]
	\centering
	\caption{MAE of different online methods with horizons in \{24, 48, 96\}. The best results are marked in \textbf{bold}.
 }\label{tab:main_result_mae}
  \resizebox{0.9\textwidth}{!}{
  \begin{tabular}{cc|ccc|ccc|ccc|ccc|ccc}
\toprule
 & \multirow{2}{*}{\diagbox[width=2cm,height=0.85cm,innerrightsep=2pt,innerwidth=1.5cm]{\textbf{\small Method}}{\small \textbf{Dataset}\\$H$}} & \multicolumn{3}{c|}{\textbf{ETTh2}} & \multicolumn{3}{c|}{\textbf{ETTm1}} & \multicolumn{3}{c|}{\textbf{Weather}} & \multicolumn{3}{c|}{\textbf{ECL}} & \multicolumn{3}{c}{\textbf{Traffic}} \\
\multicolumn{1}{c}{\textbf{Model}} &  & \textbf{24} & \textbf{48} & \textbf{96} & \textbf{24} & \textbf{48} & \textbf{96} & \textbf{24} & \textbf{48} & \textbf{96} & \textbf{24} & \textbf{48} & \textbf{96} & \textbf{24} & \textbf{48} & \textbf{96} \\ \midrule
\multirow{6}{*}{TCN} & \textbackslash{} & 0.729 & 0.800 & 0.921 & 0.771 & 0.819 & 0.782 & 0.819 & 1.909 & 1.453 & 0.659 & 0.676 & 0.693 & 0.351 & 0.392 & 0.404 \\
 & GD & 0.727 & 0.824 & 0.946 & 0.635 & 0.674 & 0.667 & 0.406 & 0.561 & 0.703 & 0.401 & 0.446 & 0.488 & 0.320 & 0.350 & 0.383 \\
 & FSNet & 0.705 & 0.829 & 0.957 & 0.521 & 0.581 & 0.609 & 0.406 & 0.558 & 0.690 & 0.390 & 0.436 & 0.474 & 0.325 & 0.355 & 0.384 \\
 & OneNet & 0.728 & 0.902 & 1.096 & 0.688 & 0.781 & 0.678 & 0.456 & 0.591 & 0.707 & 0.767 & 0.721 & 0.505 & 0.319 & 0.366 & 0.393 \\
 & SOLID++ & 0.711 & 0.799 & 0.916 & 0.524 & 0.566 & 0.590 & 0.398 & 0.543 & 0.667 & 0.394 & \textbf{0.430} & 0.472 & 0.308 & 0.334 & 0.353 \\
 & \modelns & \textbf{0.659} & \textbf{0.767} & \textbf{0.890} & \textbf{0.447} & \textbf{0.521} & \textbf{0.553} & \textbf{0.382} & \textbf{0.493} & \textbf{0.637} & \textbf{0.387} & 0.431 & \textbf{0.463} & \textbf{0.291} & \textbf{0.308} & \textbf{0.332} \\ \midrule
\multirow{5}{*}{PatchTST} & \textbackslash{} & 0.684 & 0.770 & 0.891 & 0.450 & 0.528 & 0.550 & 0.373 & 0.491 & 0.612 & 0.297 & 0.321 & 0.346 & 0.277 & 0.276 & 0.286 \\
 & GD & 0.611 & 0.718 & 0.864 & 0.407 & 0.474 & 0.509 & 0.369 & 0.482 & 0.592 & 0.294 & 0.315 & 0.341 & 0.276 & 0.267 & 0.278 \\
 & OneNet & 0.631 & 0.731 & 0.862 & 0.477 & 0.504 & 0.541 & 0.392 & 0.496 & 0.605 & 0.291 & 0.314 & 0.340 & 0.259 & 0.259 & 0.263 \\
 & SOLID++ & 0.670 & 0.761 & 0.890 & 0.406 & 0.467 & \textbf{0.504} & 0.372 & 0.485 & 0.597 & 0.294 & 0.315 & 0.340 & 0.276 & 0.264 & 0.277 \\
 & \modelns & \textbf{0.579} & \textbf{0.692} & \textbf{0.849} & \textbf{0.392} & \textbf{0.463} & 0.505 & \textbf{0.367} & \textbf{0.477} & \textbf{0.591} & \textbf{0.283} & \textbf{0.307} & \textbf{0.339} & \textbf{0.233} & \textbf{0.244} & \textbf{0.249} \\ \midrule
\multirow{5}{*}{iTransformer} & \textbackslash{} & 0.724 & 0.822 & 0.944 & 0.478 & 0.566 & 0.584 & 0.473 & 0.581 & 0.676 & 0.289 & 0.315 & 0.344 & 0.257 & 0.267 & 0.280 \\
 & GD & 0.669 & 0.788 & 0.943 & 0.422 & 0.488 & 0.529 & 0.444 & 0.552 & 0.651 & 0.287 & 0.313 & 0.341 & 0.251 & 0.262 & 0.272 \\
 & OneNet & 0.663 & 0.775 & 0.916 & 0.449 & 0.525 & 0.546 & 0.411 & 0.526 & 0.633 & 0.289 & 0.316 & 0.345 & 0.252 & 0.264 & 0.277 \\
 & SOLID++ & 0.673 & 0.789 & 0.929 & 0.421 & 0.488 & 0.527 & 0.447 & 0.554 & 0.653 & 0.288 & 0.314 & 0.343 & 0.248 & 0.260 & 0.269 \\
 & \modelns & \textbf{0.633} & \textbf{0.753} & \textbf{0.889} & \textbf{0.398} & \textbf{0.461} & \textbf{0.500} & \textbf{0.378} & \textbf{0.495} & \textbf{0.602} & \textbf{0.281} & \textbf{0.306} & \textbf{0.331} & \textbf{0.231} & \textbf{0.243} & \textbf{0.254}\\ \bottomrule
\end{tabular}
}
\end{table*}

In this section, we present experiments on real-world datasets to evaluate the effectiveness and efficiency of \modelns.

\subsection{Experimental Settings}

\noindentparagraph{\textbf{\emph{{Datasets.}}}}
 As introduced in Sec.~\ref{sec:analysis:datasets}, we use five popular benchmarks, of which more details are provided in Appendix~\ref{sec:datasets}. We conform to the dataset split ratio of FSNet and OneNet, \textit{i.e.}, we split the datasets by 20:5:75 for training, validation, and testing. The rationale is that online learning is of great practical value in scenarios of limited training data, when pretrained models are inadequate at handling new concepts during long-term online service.

\noindentparagraph{\textbf{\emph{{Forecast Models.}}}}
As our framework is model-agnostic, we pretrain three popular and advanced forecasting models, including TCN~\cite{TCN}, PatchTST~\cite{PatchTST}, and iTransformer~\cite{iTransformer}. We report the forecasting errors of these pretrained models without any online learning method (Method=" \textbackslash{}" in Table~\ref{tab:main_result_mse}-\ref{tab:main_result_mae}). 

\noindentparagraph{\textbf{\emph{{Online Learning Baselines.}}}}
We compare \model with GD, a na\"ive online gradient descent method, and the state-of-the-art online model adaptation methods, including FSNet, OneNet, and SOLID. We reimplement FSNet and OneNet by adopting the \textit{Practical} strategy described in Sec.~\ref{sec:variants}. Furthermore, FSNet is specially designed for TCN, while OneNet is based on online ensembling and is model-agnostic. We implement multiple variants of OneNet by combining each forecast model with an FSNet. As for SOLID, its original work learns linear probing with the pretrained parameters at each update time and does not inherit the fine-tuned parameters from the last update. As our datasets have a much longer online phase, we implement a variant called SOLID++ that continually fine-tunes all model parameters across the online data. Empirically, SOLID++ performs better than SOLID in our evaluation setting. 

\subsection{Overall Comparison}
\subsubsection{Effectiveness}
To verify the effectiveness of our proposed framework, we compare the predictive performance of \model and other baselines on the five popular datasets. We repeat each experiment with 3 runs and report the average results. As shown in Table~\ref{tab:main_result_mse}-\ref{tab:main_result_mae}, \model achieves the best performance in most cases, reducing the average forecast errors of all models without online learning by 42.3\%, 10.3\%, 12.9\% for TCN, PatchTST, and iTransformer, respectively. Also, it is worth mentioning that all forecast models have been enhanced by RevIN~\cite{RevIN}, a data adaptation approach to concept drift by reducing the distribution shifts w.r.t. the mean and standard deviation of time series. Since all online learning methods can still achieve remarkable improvements, it suffices to support our claim that time series forecast models need model adaptation to handle more complex concept drifts. 

Based on the same forecast model, \model outperforms the existing online model adaptation methods by 12.5\%, 13.6\%, 6.7\% against FSNet, OneNet, and SOLID++, respectively. In particular, we can observe that TCN with existing online learning methods still lags behind a frozen PatchTST. By contrast, TCN enhanced by \model can outperform the frozen PatchTST and iTransformer in some cases of ETTh2, ETTm1, and Weather datasets. Note that these 3 datasets have relatively more significant concept drift as shown in Table~\ref{tab:mse gap}. This indicates that \model has the capability of handling the concept drift during online learning. 

\subsubsection{Efficiency}
\begin{figure}[t]
	\centering
    \includegraphics[width=\linewidth]{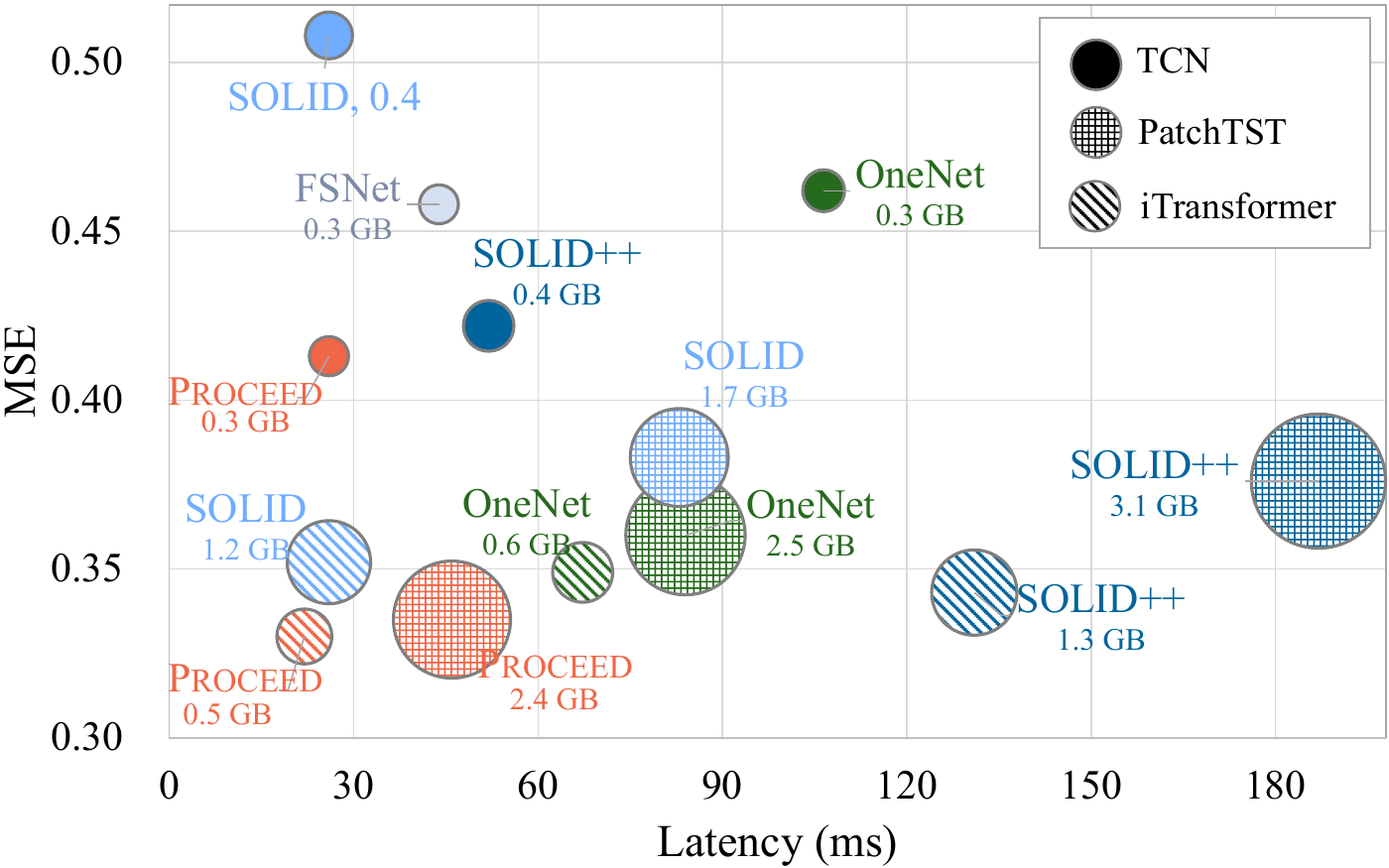}
	\caption{Efficiency comparison on the Traffic dataset ($H=24$). The horizontal axis is the online forecasting latency (millisecond) between updating the model and obtaining online predictions. The vertical axis is the average MSE on test data. The size of each circle represents the peak amount of GPU memory occupation (GB).}
  \label{fig:efficiency}
\end{figure}

As depicted in Fig.~\ref{fig:efficiency}, we compare the GPU memory occupation and inference latency of different model adaptation methods on Traffic, the largest time series dataset. We record the online latency, including feedback-based model adaptation and any other additional steps. Here, SOLID only fine-tunes the final layer rather than the whole model, reducing the gradient computation cost. More specifically, SOLID and SOLID++ compute gradients on several selected training samples iteratively instead of concurrently, so as to save GPU memory. By contrast, \model only uses the latest training sample and thereby achieves the lowest latency. Moreover, \model is faster and more lightweight than the online ensembling method OneNet.

\subsection{Visualization of the Representation Space}\label{sec:visual}

\begin{figure}[t]
  \centering
  \subcaptionbox{\label{fig:visual_concept}Concepts}
  {
    \includegraphics[width=.47\linewidth]{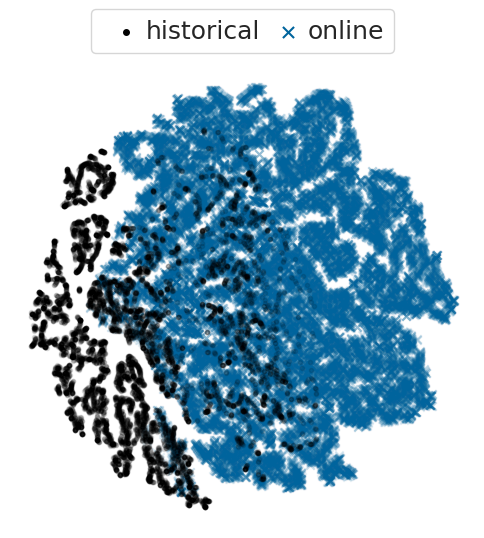}
  }
  \subcaptionbox{\label{fig:visual_drift}Concept drifts}
  {
    \includegraphics[width=.47\linewidth]{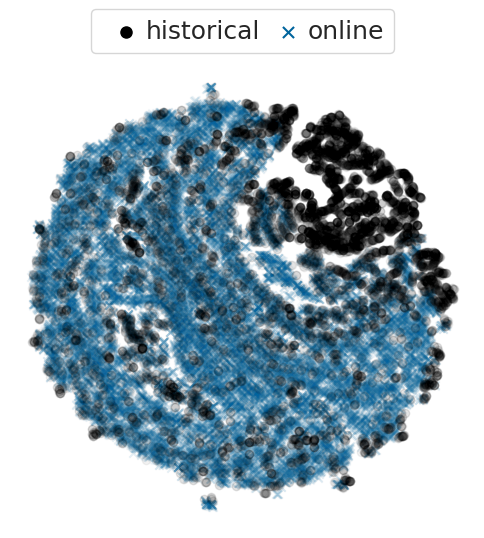}
  }
  \caption{Visualization of concept representations and concept drift representations on ETTm1 ($H=24$).
  }
  \label{fig:visual}
\end{figure}

To verify our assumption that online data have OOD concepts, we adopt t-SNE to visualize the representations of concepts $\mathbf c_{t}$ and concept drifts $\boldsymbol{\delta}_{t-H\to t}$ on the historical training data and online data. As shown in Fig.~\ref{fig:visual_concept}, there are a great number of OOD concepts in the online data, which are distinct from any historical concept. By contrast, as shown in Fig.~\ref{fig:visual_drift}, we encounter much fewer OOD concept drifts on the online data. 
These results support our intuition illustrated in Fig.~\ref{fig:vs}. Hence, it is desirable for the adaptation generator to generate adaptations based on concept drifts instead of concepts.

\subsection{Ablation Study}\label{sec:ablation}
\begin{table}[t]
	\centering
	\caption{Ablation study results. We report the average MSE with horizons in $\{24, 48, 96\}$ and the relative MSE increase ($\bar\Delta_{\text{MSE}}$) of the variants against \model on each dataset.
 }\label{tab:ablation}
  \resizebox{\linewidth}{!}{
\begin{tabular}{cc|ccccc|c}
\toprule
Model                         & Method                  & ETTh2          & ETTm1          & Weather        & ECL            & Traffic & $\bar\Delta_{\text{MSE}}$ \\ \midrule
\multirow{5}{*}{\begin{sideways}PatchTST\end{sideways}}     & feedback-only & 3.765 & 0.576 & 0.992 & 4.899 & 0.385 & 4.9\% \\
&$\mathcal G(\mathbf c_{t})$ & 4.194 & 0.607 & 0.997 & 4.742 & 0.358 & 6.5\% \\
&$\mathcal E'(\mathbf X_{t-H})$ & 3.702 & 0.678 & 1.011 & 4.974 & 0.363 & 7.8\% \\
&\textit{diff. }$\mathbf W_1^{(\ell)},\mathbf W_2^{(\ell)}$ & 3.573 & 0.578 & 0.993 & 4.765 & 0.358 & 1.9\% \\
&\modelns & \textbf{3.468} & \textbf{0.554} & \textbf{0.986} & \textbf{4.732} & \textbf{0.356} & - \\ \midrule
\multirow{5}{*}{\begin{sideways}iTransformer\end{sideways}} & feedback-only & 4.506 & 0.590 & 1.140 & 4.965 & 0.368 & 7.3\% \\
&$\mathcal G(\mathbf c_{t})$ & 4.218 & 0.558 & 1.030 & 4.842 & \textbf{0.352} & 1.2\% \\
&$\mathcal E'(\mathbf X_{t-H})$ & 4.396 & 0.557 & 1.030 & 4.776 & \textbf{0.352} & 1.7\% \\
&\textit{diff. }$\mathbf W_1^{(\ell)},\mathbf W_2^{(\ell)}$ & 4.254 & 0.547 & 1.020 & 4.767 & \textbf{0.352} & 0.4\% \\
&\modelns & \textbf{4.216} & \textbf{0.543} & \textbf{1.017} & \textbf{4.752} & \textbf{0.352} & - \\ \bottomrule
\end{tabular}
}
\end{table}
For a fine-grained investigation into the role of our proposed components, we conduct more ablation studies by introducing five variants as follows. (1) \textit{feedback-only}: this variant only performs gradient descent based on feedback from $\mathcal D_{t-}$; (2) $\mathcal G(\mathbf c_{t})$: this variant generates adaptation based on the concept of the test sample only instead of estimating concept drift; (3) $\mathcal E'(\mathbf X_{t-H})$: this variant uses the same encoder $\mathcal E'$ to extract concepts $\mathbf c_{t-H}$ and $\mathbf c_t$ from lookback windows $\mathbf X_{t-H}$ and $\mathbf X_{t}$, respectively; (4) \textit{diff. } $\mathbf W_1^{(\ell)},\mathbf W_2^{(\ell)}$: the bottleneck layers that generate adaptation coefficients are totally different across the model layers.

As shown in Table~\ref{tab:ablation}, we have four major observations. First, \model outperforms the na\"ive feedback-based model adaptation method by a large margin, demonstrating the necessity of proactive model adaptation in reducing the distribution gap between new training data and test data. Second, the variant \textit{w/ } $\mathcal G(\mathbf c_{t})$, which generates adaptation only based on the estimated test concept, results in a considerably higher MSE than \model based on concept drift. This is due to the fact that some OOD concepts may take place in the online phase and challenge the adaptation generator. By contrast, our concept drift-based design can make the inputs of $\mathcal G$ in-distribution as much as possible, reducing difficulties in generating adaptation.
Third, the variant \textit{w/ } $\mathcal E'(\mathbf X_{t-H})$ has suboptimal performance compared with \model with two different concept encoders. One reason is that we can only infer limited information about $\mathbf Y_{t-H}$ from $\mathbf X_{t-H}$, and it would be better to leverage the observed $\mathbf Y_{t-H}$. Another possible reason is that the model before proactive model adaptation does not overfit $\mathcal D_{t-}$, while we anticipate the adapted model being optimal for $(\mathbf X_t, \mathbf Y_t)$. Thus, $\mathcal E$ should encode a little information of the latest training sample into $\mathbf c_{t-H}$, while a different encoder $\mathcal E'$ should encode all information of the test sample into $\mathbf c_{t}$. 
Fourth, the variant learning different $\mathbf W_1^{(\ell)}$ and $\mathbf W_2^{(\ell)}$ do not significantly outperform \model with weights shared across some layers. Considering the non-stationarity of time series, an over-parameterized generator $\mathcal G$ has higher overfitting risks on training data. Thus we only set the bias term $\mathbf b_1^{(\ell)}$ to be different across model layers, yielding distinct, layer-specific adaptations.


Besides, we study different kinds of operations on multivariate time series using our concept encoder. Let $\mathbf c_{t}^{(i)}$ denote the latent concept feature vector of the $i$-th variate. Apart from the average introduced in Eq.~(\ref{eq:avg}), we implement two variants as follows:
\begin{itemize}[leftmargin=*]
    \item \textbf{linear transformation}: we learn a linear layer that transforms the concatenation of all latent features, i.e., $c_t=W^\top[c_{t}^{(0)},c_{t}^{(1)},\cdots]$.
    \item \textbf{weighted sum}: we learn an individual weight $w_i\in \mathbb R$ for each $i$-th variate, and $c_t=\sum_i w_ic_{t}^{(i)}$. 
\end{itemize}
In Table~\ref{tab:encoder}, we compare the performance of the variants in terms of the average MSE with $H\in \{24,48,96\}$. The results demonstrate that the average operation is simple yet effective. One possible reason is that all variates are typically assumed to have equal importance during both training and evaluation. 

\begin{table}[t]
	\centering
	\caption{MSE of \model with various concept encoders.
 }\label{tab:encoder}
  \resizebox{\linewidth}{!}{
\begin{tabular}{cc|ccccc}
\toprule
Model                         & Method                  & ETTh2          & ETTm1          & Weather        & ECL            & Traffic \\ \midrule
\multirow{3}{*}{PatchTST}   &  linear trans. & 3.498 & 0.570 & \textbf{0.985} & 5.661 & 0.363 \\
&weighted sum & \textbf{3.467} & 0.571 & 0.986 & 5.435 & 0.362 \\
&\textbf{average} & 3.468 & \textbf{0.554} & 0.986 & \textbf{4.732} & \textbf{0.356} \\\midrule
\multirow{3}{*}{iTransfo.} & linear trans. & 4.400 & 0.548 & 1.018 & 4.944 & 0.355 \\
&weighted sum & 4.441 & 0.551 & 1.021 & 5.260 & \textbf{0.352} \\
&\textbf{average} & \textbf{4.216} & \textbf{0.543} & \textbf{1.017} & \textbf{4.752} & \textbf{0.352} \\ \bottomrule
\end{tabular}
}
\end{table}

\section{Conclusion}
In this work, we highlight that online time series forecasting inherently has a temporal gap between each test sample and available training data, where concept drift may well occur. Through empirical study, we found that this gap hinders the effectiveness of existing model adaptation methods which passively rely on the feedback in forecasting recent data. To address this issue, we propose a novel online model adaptation framework named \modelns, which proactively adapts the forecast model to the test concept before forecasting each test sample. Specifically, \model first fine-tunes the model on the latest acquired training sample, then extracts latent features from time series data to estimate the undergoing concept drift, and efficiently maps the estimated drift into parameter adjustments that are tailored for the test sample. Furthermore, we synthesize diverse concept drift and optimize \model with the generalization capacity of mapping concept drift to beneficial parameter adjustments. Extensive experiments on five real-world time series datasets demonstrate that our proposed \model remarkably reduces the forecast errors of various forecast models and surpasses the state-of-the-art online learning methods. Our proactive model adaptation method provides a new direction in addressing continuous concept drift for online systems, which we wish to help any other prediction tasks with a feedback delay issue.

\begin{acks}
This work is supported by the National Key Research and Development Program of China  (2022YFE0200500), Shanghai Municipal Science and Technology Major Project  (2021SHZDZX0102), and SJTU Global Strategic Partnership Fund (2021 SJTU-HKUST). We would like to thank Shaofeng Cai for his valuable advice on the preliminary version of the paper.
\end{acks}

\bibliographystyle{ACM-Reference-Format}
\balance
\bibliography{main.bib}

@InProceedings{NSTransformer,
  author       = {Liu, Yong and Wu, Haixu and Wang, Jianmin and Long, Mingsheng},
  booktitle    = {Advances in Neural Information Processing Systems},
  title        = {Non-stationary Transformers: Exploring the Stationarity in Time Series Forecasting},
  year         = {2022},
  pages        = {9881--9893},
  volume       = {35},
  creationdate = {2024-08-06T21:05:13},
}

@article{Sun2023DSE,
  title = {Graph Neural Network-Based Short‑Term Load Forecasting with Temporal Convolution},
  volume = {9},
  ISSN = {2364-1541},
  url = {http://dx.doi.org/10.1007/s41019-023-00233-8},
  DOI = {10.1007/s41019-023-00233-8},
  number = {2},
  journal = {Data Science and Engineering},
  publisher = {Springer Science and Business Media LLC},
  author = {Sun,  Chenchen and Ning,  Yan and Shen,  Derong and Nie,  Tiezheng},
  year = {2023},
  month = nov,
  pages = {113–132}
}

@InProceedings{MOIRAI,
  author       = {Woo, Gerald and Liu, Chenghao and Kumar, Akshat and Xiong, Caiming and Savarese, Silvio and Sahoo, Doyen},
  booktitle    = {Proceedings of the 41st International Conference on Machine Learning},
  title        = {Unified Training of Universal Time Series Forecasting Transformers},
  year         = {2024},
  editor       = {Salakhutdinov, Ruslan and Kolter, Zico and Heller, Katherine and Weller, Adrian and Oliver, Nuria and Scarlett, Jonathan and Berkenkamp, Felix},
  month        = {21--27 Jul},
  pages        = {53140--53164},
  publisher    = {PMLR},
  series       = {Proceedings of Machine Learning Research},
  volume       = {235},
  creationdate = {2024-09-04T12:32:09},
}

@inproceedings{
SOFTS,
title={{SOFTS}: Efficient Multivariate Time Series Forecasting with Series-Core Fusion},
author={Lu Han and Xu-Yang Chen and Han-Jia Ye and De-Chuan Zhan},
booktitle={The Thirty-eighth Annual Conference on Neural Information Processing Systems},
year={2024},
}

@InProceedings{DoubleAdapt,
  author       = {Zhao, Lifan and Kong, Shuming and Shen, Yanyan},
  booktitle    = {Proceedings of the 29th ACM SIGKDD Conference on Knowledge Discovery and Data Mining},
  title        = {DoubleAdapt: A Meta-learning Approach to Incremental Learning for Stock Trend Forecasting},
  year         = {2023},
  address      = {New York, NY, USA},
  pages        = {3492–3503},
  publisher    = {Association for Computing Machinery},
  series       = {KDD '23},
  creationdate = {2024-01-26T22:28:29},
  doi          = {10.1145/3580305.3599315},
  isbn         = {9798400701030},
  location     = {Long Beach, CA, USA},
  numpages     = {12},
  url          = {https://doi.org/10.1145/3580305.3599315},
}

@InProceedings{LLF,
  author    = {Xiaoyu You and Mi Zhang and Daizong Ding and Fuli Feng and Yuanmin Huang},
  booktitle = {Proceedings of the 30th {ACM} International Conference on Information {\&} Knowledge Management},
  title     = {Learning to Learn the Future: Modeling Concept Drift in Time Series Prediction},
  year      = {2021},
  month     = {oct},
  publisher = {{ACM}},
  doi       = {10.1145/3459637.3482271},
}

@Article{DDGDA,
  author       = {Wendi Li and Xiao Yang and Weiqing Liu and Yingce Xia and Jiang Bian},
  title        = {{DDG}-{DA}: Data Distribution Generation for Predictable Concept Drift Adaptation},
  year         = {2022},
  month        = {jun},
  number       = {4},
  pages        = {4092--4100},
  publisher    = {Association for the Advancement of Artificial Intelligence ({AAAI})},
  volume       = {36},
  creationdate = {2022-10-27T18:41:04},
  doi          = {10.1609/aaai.v36i4.20327},
  journal      = {Proceedings of the {AAAI} Conference on Artificial Intelligence},
}

@InProceedings{OSAKA,
  author    = {Massimo Caccia and Pau Rodríguez and Oleksiy Ostapenko and Fabrice Normandin and Min Lin and Lucas Page-Caccia and Issam Hadj Laradji and Irina Rish and Alexandre Lacoste and David Vázquez and Laurent Charlin},
  booktitle = {NeurIPS},
  title     = {Online Fast Adaptation and Knowledge Accumulation (OSAKA): a New Approach to Continual Learning},
  year      = {2020},
  cdate     = {1577836800000},
}

@inproceedings{
FSNet,
title={Learning Fast and Slow for Online Time Series Forecasting},
author={Quang Pham and Chenghao Liu and Doyen Sahoo and Steven Hoi},
booktitle={The Eleventh International Conference on Learning Representations },
year={2023},
url={https://openreview.net/forum?id=q-PbpHD3EOk}
}

@InProceedings{OneNet,
  author       = {YiFan Zhang and Qingsong Wen and Xue Wang and Weiqi Chen and Liang Sun and Zhang Zhang and Liang Wang and Rong Jin and Tieniu Tan},
  booktitle    = {Thirty-seventh Conference on Neural Information Processing Systems},
  title        = {OneNet: Enhancing Time Series Forecasting Models under Concept Drift by Online Ensembling},
  year         = {2023},
  creationdate = {2024-01-29T13:51:09},
  url          = {https://openreview.net/forum?id=Q25wMXsaeZ},
}

@Article{SurveyConcepDrift,
  author       = {Jo{\~{a}}o Gama and Indr{\.{e}} {\v{Z}}liobait{\.{e}} and Albert Bifet and Mykola Pechenizkiy and Abdelhamid Bouchachia},
  journal      = {{ACM} Computing Surveys},
  title        = {A Survey on Concept Drift Adaptation},
  year         = {2014},
  month        = {apr},
  number       = {4},
  pages        = {1--37},
  volume       = {46},
  creationdate = {2022-11-21T13:46:57},
  doi          = {10.1145/2523813},
  publisher    = {Association for Computing Machinery ({ACM})},
}

@InProceedings{AdaRNN,
  author    = {Yuntao Du and Jindong Wang and Wenjie Feng and Sinno Pan and Tao Qin and Renjun Xu and Chongjun Wang},
  booktitle = {Proceedings of the 30th {ACM} International Conference on Information {\&} Knowledge Management},
  title     = {{AdaRNN}: Adaptive Learning and Forecasting for Time Series},
  year      = {2021},
  month     = {oct},
  publisher = {{ACM}},
  doi       = {10.1145/3459637.3482315},
}

@inproceedings{DER,
 author = {Buzzega, Pietro and Boschini, Matteo and Porrello, Angelo and Abati, Davide and CALDERARA, SIMONE},
 booktitle = {Advances in Neural Information Processing Systems},
 pages = {15920--15930},
 publisher = {Curran Associates, Inc.},
 title = {Dark Experience for General Continual Learning: a Strong, Simple Baseline},
 volume = {33},
 year = {2020}
}

@InProceedings{EWC,
  author       = {Schwarz, Jonathan and Czarnecki, Wojciech and Luketina, Jelena and Grabska-Barwinska, Agnieszka and Teh, Yee Whye and Pascanu, Razvan and Hadsell, Raia},
  booktitle    = {Proceedings of the 35th International Conference on Machine Learning},
  title        = {Progress \& Compress: A scalable framework for continual learning},
  year         = {2018},
  month        = {10--15 Jul},
  pages        = {4528--4537},
  publisher    = {PMLR},
  series       = {Proceedings of Machine Learning Research},
  volume       = {80},
  creationdate = {2024-04-29T13:25:45},
}

@inproceedings{SOLID,
author = {Chen, Mouxiang and Shen, Lefei and Fu, Han and Li, Zhuo and Sun, Jianling and Liu, Chenghao},
title = {Calibration of Time-Series Forecasting: Detecting and Adapting Context-Driven Distribution Shift},
year = {2024},
isbn = {9798400704901},
publisher = {Association for Computing Machinery},
address = {New York, NY, USA},
url = {https://doi.org/10.1145/3637528.3671926},
doi = {10.1145/3637528.3671926},
booktitle = {Proceedings of the 30th ACM SIGKDD Conference on Knowledge Discovery and Data Mining},
pages = {341–352},
numpages = {12},
location = {Barcelona, Spain},
series = {KDD '24}
}

@Article{Wang2023,
  author        = {Liyuan Wang and Xingxing Zhang and Hang Su and Jun Zhu},
  title         = {A Comprehensive Survey of Continual Learning: Theory, Method and Application},
  year          = {2023},
  month         = jan,
  archiveprefix = {arXiv},
  creationdate  = {2023-11-27T22:03:08},
  eprint        = {2302.00487},
  primaryclass  = {cs.LG},
}

@InCollection{Catastrophic,
  author    = {Michael McCloskey and Neal J. Cohen},
  booktitle = {Psychology of Learning and Motivation},
  publisher = {Elsevier},
  title     = {Catastrophic Interference in Connectionist Networks: The Sequential Learning Problem},
  year      = {1989},
  pages     = {109--165},
  doi       = {10.1016/s0079-7421(08)60536-8},
}

@InProceedings{DRAIN,
  author       = {Guangji Bai and Chen Ling and Liang Zhao},
  booktitle    = {The Eleventh International Conference on Learning Representations},
  title        = {Temporal Domain Generalization with Drift-Aware Dynamic Neural Networks},
  year         = {2023},
  creationdate = {2023-11-06T15:01:02},
  url          = {https://openreview.net/forum?id=sWOsRj4nT1n},
}

@Article{DLinear,
  author       = {Ailing Zeng and Muxi Chen and Lei Zhang and Qiang Xu},
  journal      = {Proceedings of the {AAAI} Conference on Artificial Intelligence},
  title        = {Are Transformers Effective for Time Series Forecasting?},
  year         = {2023},
  month        = {jun},
  number       = {9},
  pages        = {11121--11128},
  volume       = {37},
  creationdate = {2023-07-18T20:35:56},
  doi          = {10.1609/aaai.v37i9.26317},
  publisher    = {Association for the Advancement of Artificial Intelligence ({AAAI})},
}

@Article{Hoi2021,
  author       = {Hoi, Steven C.H. and Sahoo, Doyen and Lu, Jing and Zhao, Peilin},
  journal      = {Neurocomputing},
  title        = {Online learning: A comprehensive survey},
  year         = {2021},
  issn         = {0925-2312},
  month        = oct,
  pages        = {249--289},
  volume       = {459},
  creationdate = {2024-02-05T13:46:54},
  doi          = {10.1016/j.neucom.2021.04.112},
  publisher    = {Elsevier BV},
}

@Article{Boese2017,
  author       = {Böse, Joos-Hendrik and Flunkert, Valentin and Gasthaus, Jan and Januschowski, Tim and Lange, Dustin and Salinas, David and Schelter, Sebastian and Seeger, Matthias and Wang, Yuyang},
  journal      = {Proceedings of the VLDB Endowment},
  title        = {Probabilistic demand forecasting at scale},
  year         = {2017},
  issn         = {2150-8097},
  month        = aug,
  number       = {12},
  pages        = {1694--1705},
  volume       = {10},
  creationdate = {2024-04-21T20:22:44},
  doi          = {10.14778/3137765.3137775},
  publisher    = {Association for Computing Machinery (ACM)},
}

@InProceedings{SolarNIPS,
 author = {Boussif, Oussama and Boukachab, Ghait and Assouline, Dan and Massaroli, Stefano and Yuan, Tianle and Benabbou, Loubna and Bengio, Yoshua},
 booktitle = {Advances in Neural Information Processing Systems},
 pages = {2342--2367},
 publisher = {Curran Associates, Inc.},
 title = {Improving day-ahead  Solar Irradiance Time Series Forecasting by Leveraging Spatio-Temporal Context},
 volume = {36},
 year = {2023},
  creationdate = {2024-04-21T22:09:25},
}

@Article{METER,
  author       = {Zhu, Jiaqi and Cai, Shaofeng and Deng, Fang and Ooi, Beng Chin and Zhang, Wenqiao},
  journal      = {Proceedings of the VLDB Endowment},
  title        = {METER: A Dynamic Concept Adaptation Framework for Online Anomaly Detection},
  year         = {2023},
  issn         = {2150-8097},
  month        = dec,
  number       = {4},
  pages        = {794--807},
  volume       = {17},
  creationdate = {2024-04-21T18:56:43},
  doi          = {10.14778/3636218.3636233},
  publisher    = {Association for Computing Machinery (ACM)},
}

@InProceedings{GPT4TS,
  author       = {Tian Zhou and Peisong Niu and Xue Wang and Liang Sun and Rong Jin},
  booktitle    = {Thirty-seventh Conference on Neural Information Processing Systems},
  title        = {One Fits All: Power General Time Series Analysis by Pretrained {LM}},
  year         = {2023},
  creationdate = {2023-11-06T15:12:30},
  url          = {https://openreview.net/forum?id=gMS6FVZvmF},
}

@InProceedings{PatchTST,
  author       = {Yuqi Nie and Nam H Nguyen and Phanwadee Sinthong and Jayant Kalagnanam},
  booktitle    = {The Eleventh International Conference on Learning Representations},
  title        = {A Time Series is Worth 64 Words: Long-term Forecasting with Transformers},
  year         = {2023},
  creationdate = {2023-03-10T16:09:54},
  url          = {https://openreview.net/forum?id=Jbdc0vTOcol},
}

@InProceedings{MTGNN,
  author       = {Zonghan Wu and Shirui Pan and Guodong Long and Jing Jiang and Xiaojun Chang and Chengqi Zhang},
  booktitle    = {Proceedings of the 26th {ACM} {SIGKDD} International Conference on Knowledge Discovery \& Data Mining},
  title        = {Connecting the Dots: Multivariate Time Series Forecasting with Graph Neural Networks},
  year         = {2020},
  month        = {aug},
  publisher    = {{ACM}},
  creationdate = {2022-07-27T09:36:28},
  doi          = {10.1145/3394486.3403118},
}

@Article{Informer,
  author       = {Haoyi Zhou and Shanghang Zhang and Jieqi Peng and Shuai Zhang and Jianxin Li and Hui Xiong and Wancai Zhang},
  journal      = {Proceedings of the {AAAI} Conference on Artificial Intelligence},
  title        = {Informer: Beyond Efficient Transformer for Long Sequence Time-Series Forecasting},
  year         = {2021},
  month        = {may},
  number       = {12},
  pages        = {11106--11115},
  volume       = {35},
  creationdate = {2023-07-18T20:40:53},
  doi          = {10.1609/aaai.v35i12.17325},
  publisher    = {Association for the Advancement of Artificial Intelligence ({AAAI})},
}

@InProceedings{RevIN,
  author       = {Taesung Kim and Jinhee Kim and Yunwon Tae and Cheonbok Park and Jang-Ho Choi and Jaegul Choo},
  booktitle    = {International Conference on Learning Representations},
  title        = {Reversible Instance Normalization for Accurate Time-Series Forecasting against Distribution Shift},
  year         = {2022},
  creationdate = {2022-10-27T18:45:27},
  url          = {https://openreview.net/forum?id=cGDAkQo1C0p},
}

@inproceedings{
iTransformer,
title={iTransformer: Inverted Transformers Are Effective for Time Series Forecasting},
author={Yong Liu and Tengge Hu and Haoran Zhang and Haixu Wu and Shiyu Wang and Lintao Ma and Mingsheng Long},
booktitle={The Twelfth International Conference on Learning Representations},
year={2024},
url={https://openreview.net/forum?id=JePfAI8fah}
}

@inproceedings{SAN,
title={Adaptive Normalization for Non-stationary Time Series Forecasting: A Temporal Slice Perspective},
author={Zhiding Liu and Mingyue Cheng and Zhi Li and Zhenya Huang and Qi Liu and Yanhu Xie and Enhong Chen},
booktitle={Thirty-seventh Conference on Neural Information Processing Systems},
year={2023},
url={https://openreview.net/forum?id=5BqDSw8r5j}
}

@inproceedings{Dish-TS,
  title={Dish-TS: A General Paradigm for Alleviating Distribution Shift in Time Series Forecasting},
  author={Wei Fan and Pengyang Wang and Dongkun Wang and Dongjie Wang and Yuanchun Zhou and Yanjie Fu},
  booktitle={AAAI Conference on Artificial Intelligence},
  year={2023},
  url={https://api.semanticscholar.org/CorpusID:257232506}
}

@inproceedings{
LIFT,
title={Rethinking Channel Dependence for Multivariate Time Series Forecasting: Learning from Leading Indicators},
author={Lifan Zhao and Yanyan Shen},
booktitle={The Twelfth International Conference on Learning Representations},
year={2024},
url={https://openreview.net/forum?id=JiTVtCUOpS}
}

@Article{TCN,
  author        = {Shaojie Bai and J. Zico Kolter and Vladlen Koltun},
  title         = {An Empirical Evaluation of Generic Convolutional and Recurrent Networks for Sequence Modeling},
  year          = {2018},
  month         = mar,
  archiveprefix = {arXiv},
  eprint        = {1803.01271},
  primaryclass  = {cs.LG},
}

@Article{Pangu,
  author       = {Bi, Kaifeng and Xie, Lingxi and Zhang, Hengheng and Chen, Xin and Gu, Xiaotao and Tian, Qi},
  journal      = {Nature},
  title        = {Accurate medium-range global weather forecasting with 3D neural networks},
  year         = {2023},
  issn         = {1476-4687},
  number       = {7970},
  pages        = {533--538},
  volume       = {619},
  creationdate = {2024-12-05T19:38:04},
  doi          = {10.1038/s41586-023-06185-3},
  publisher    = {Springer Science and Business Media LLC},
}

@inproceedings{
DSOF,
title={Fast and Slow Streams for Online Time Series Forecasting Without Information Leakage},
author={Ying-yee Ava Lau and Zhiwen Shao and Dit-Yan Yeung},
booktitle={The Thirteenth International Conference on Learning Representations},
year={2025},
url={https://openreview.net/forum?id=I0n3EyogMi},
}

@misc{ActNow,
      title={Act Now: A Novel Online Forecasting Framework for Large-Scale Streaming Data}, 
      author={Daojun Liang and Haixia Zhang and Jing Wang and Dongfeng Yuan and Minggao Zhang},
      year={2024},
      eprint={2412.00108},
      archivePrefix={arXiv},
      primaryClass={cs.LG},
      url={https://arxiv.org/abs/2412.00108}, 
}

@article{duan2025brain,
  title={Brain-inspired online adaptation for remote sensing with spiking neural network},
  author={Duan, Dexin and Liu, Peilin and Hui, Bingwei and Wen, Fei},
  journal={IEEE Transactions on Geoscience and Remote Sensing},
  year={2025},
  publisher={IEEE}
}

\appendix

\section{Theoretical Analysis}\label{sec:theorem}
Inspired by a recent work~\cite{DRAIN}, we can prove that it is feasible for the forecast model with proactive model adaptation to have lower forecasting error than one without proactive model adaptation.

In practice, we initialize the parameters ${\mathbf W}_2^{(\ell)}$ as an all-zero matrix, and the adaptation coefficients stem from zeros. Thus, we have $\mathcal A(\boldsymbol{\theta}; \boldsymbol\phi_0) = \boldsymbol{\theta}$, where $\mathcal A$ denotes the model adpater and $\boldsymbol\phi_0$ denotes the initialized parameters of $\mathcal A$. 
Given randomly shuffled historical data, we train the adapter parameters into $\boldsymbol\phi$ that is close to an optimum $\boldsymbol\phi^*$ and approximates the transition probability $P(\boldsymbol{\theta}_{t} \mid \boldsymbol{\theta}_{s}, \mathbf{X}_{s}, \mathbf{X}_{t})$. Let $\hat{\boldsymbol\theta}_t$ denote $\mathcal A(\boldsymbol\theta_{t-H}; \boldsymbol\phi)$, $\boldsymbol\theta_t^*$ denote the optimal model parameters for forecasting $\mathbf{Y}_t$, and $\boldsymbol\theta_t^*=\mathcal A(\boldsymbol\theta_{t-H}; \boldsymbol\phi^*)$. The adapted model parameters $\hat{\boldsymbol{\theta}}_{t}$ generated by the well-learned model adapter are expected to get closer to the optimal model parameters $\boldsymbol\theta_t^*$ than those generated by $\mathcal A(\cdot; \boldsymbol\phi_0)$. Formally, we have
\begin{equation}\label{eq:close}
    \| \mathcal A(\boldsymbol\theta_{t-H}; \boldsymbol\phi) - \mathcal A(\boldsymbol\theta_{t-H}; \boldsymbol\phi^*) \| <  \| \mathcal A(\boldsymbol\theta_{t-H}; \boldsymbol\phi_0) - \mathcal A(\boldsymbol\theta_{t-H}; \boldsymbol\phi^*) \|,
\end{equation}
\textit{i.e.}, 
\begin{equation}\label{eq:parameter distance}
\| \hat{\boldsymbol\theta}_t - \boldsymbol\theta_t^* \| <  \| \boldsymbol\theta_{t-H} - \boldsymbol\theta_t^* \|.    
\end{equation}
As we synthesize various concept drifts to train the model adapter, we believe that most concept drifts on test data have been learned in past experience, thereby satisfying Eq.~(\ref{eq:parameter distance}) during the online phase. 

Assuming that the forecast model $\mathcal F$ has Lipschitz constant with upper bound $\mathcal L_{upper}$ and lower bound $\mathcal L_{lower}$ \textit{w.r.t.} its parameters $\boldsymbol\theta$, we have
\begin{equation}
    \mathcal L_{lower} \| \boldsymbol{\theta} - \boldsymbol{\theta}^{'} \| <  \| \mathcal F(\mathbf X; \boldsymbol\theta) - \mathcal F(\mathbf X; \boldsymbol\theta^{'}) \| < \mathcal L_{upper} \| \boldsymbol{\theta} - \boldsymbol{\theta}^{'} \|, \forall \mathbf X.
\end{equation}
Then, we can derive the following inequalities:
\begin{equation}\begin{aligned}\label{eq:two bounds}
    \| \mathcal F(\mathbf X_t; \boldsymbol\theta_{t-H}) - \mathcal F(\mathbf X_t; \boldsymbol\theta^*_t) \| &> \mathcal L_{lower} \| \boldsymbol{\theta}_{t-H} - \boldsymbol{\theta}^*_t \|, \\
    \| \mathcal F(\mathbf X_t; \hat{\boldsymbol\theta}_t) - \mathcal F(\mathbf X_t; \boldsymbol\theta^*_t) \| &< \mathcal L_{upper} \| \hat{\boldsymbol\theta}_t - \boldsymbol{\theta}^*_t \|.
\end{aligned}\end{equation}
Note that $\mathcal F$ is usually a neural network as a continuous function of $\boldsymbol{\theta}$. We define $\mathcal{S}(\boldsymbol{\theta}^*_t, \Delta)$ as a sphere centering at $\boldsymbol{\theta}^*_t$ with a radius $\Delta \in \mathbb R^+$. When $\Delta$ approaches 0, due to the continuity of $\mathcal F$, the upper bound and lower bound of Lipschitz constant within $\mathcal{S}(\boldsymbol{\theta}^*_t, \Delta)$ will become closer and finally identical, \textit{i.e.}, $\lim_{\Delta\to 0^+} \mathcal L_{upper}/\mathcal L_{lower} = 1$. Moreover, we have known that $\| \boldsymbol{\theta}_{t-H} - \boldsymbol{\theta}^*_t \| /\| \hat{\boldsymbol\theta}_t - \boldsymbol{\theta}^*_t \| > 1$ in Eq.~(\ref{eq:parameter distance}). Therefore, there exists a constant $\Delta > 0$ such that $\mathcal L_{upper}/\mathcal L_{lower} < \| \boldsymbol{\theta}_{t-H} - \boldsymbol{\theta}^*_t \| /\| \hat{\boldsymbol\theta}_t - \boldsymbol{\theta}^*_t \|$, where $\boldsymbol{\theta}_{t-H},\hat{\boldsymbol\theta}_t\in \mathcal{S}(\boldsymbol{\theta}^*_t, \Delta)$. Thus it is possible to satisfy the following inequality:
\begin{equation}\label{eq:bound}
    \mathcal L_{upper}\| \hat{\boldsymbol\theta}_t - \boldsymbol{\theta}^*_t \| < \mathcal L_{lower}\| \boldsymbol{\theta}_{t-H} - \boldsymbol{\theta}^*_t \|.
\end{equation}
Combining Eq.~(\ref{eq:bound}) and Eq.~(\ref{eq:two bounds}), we have:
\begin{equation}
    \| \mathcal F(\mathbf X_t; \hat{\boldsymbol\theta}_t) - \mathcal F(\mathbf X_t; \boldsymbol\theta^*_t) \| < \| \mathcal F(\mathbf X_t; \boldsymbol\theta_{t-H}) - \mathcal F(\mathbf X_t; \boldsymbol\theta^*_t) \|, 
\end{equation}
where $\boldsymbol\theta^*_t$ is the optimal parameters for forecasting $\mathbf X_t$. In other words, it is possible for the forecast model with proactive model adaptation that yields fewer forecast errors, \textit{i.e.},
\begin{equation}
    \| \mathcal F(\mathbf X_t; \hat{\boldsymbol\theta}_t) - \mathbf Y_t \| < \| \mathcal F(\mathbf X_t; \boldsymbol\theta_{t-H}) - \mathbf Y_t \|.
\end{equation}

\section{Related Works}

\subsection{Online Model Adaptation}
Online model adaptation, or online learning~\cite{Hoi2021}, has been a popular learning paradigm that updates models on new data instantly or periodically. In the general field, most efforts focus on addressing the catastrophic forgetting issue~\cite{Catastrophic} stemming from excessive updates, and numerous \textit{continual learning}~\cite{Wang2023} methods have been developed to retain acquired knowledge of past data by rehearsal mechanisms~\cite{DER} and regularization terms~\cite{EWC}, which can be seamlessly incorporated into our framework. 
Most recently, SOLID~\cite{SOLID} proposed to adapt the forecast model by several selected training samples which are assumed to have a similar context (\textit{i.e.}, concept in this paper) to each test sample. SOLID~\cite{SOLID} relies on heuristic measures about context similarity and is not fully aware of unobserved contexts. The selected training samples may not share the exactly same concept with the test sample. Thus it is desirable to further adopt our proposed framework. We would like to leave the combination of training data sampling and proactive model adaptation as future work.

After the submission of our manuscript, there are two concurrent researches~\cite{DSOF, ActNow} that also noticed the information leakage in FSNet and OneNet. Since the ground-truth $H$-step values of the last prediction are not fully observed, both works propose to generate pseudo labels of the unobserved values, concatenate the pseudo labels with observed labels to simulate complete ground truth, and calculate the forecast errors to update the model. The potential drawback is that the pseudo label generator is still susceptible to concept drift and may produce low-quality pseudo labels, leaving the concept drift issue unresolved.

On top of online learning, \textit{rolling retraining}~\cite{DDGDA} is another learning paradigm that periodically re-trains a new model from scratch on all historical data, while the cost of frequent retraining is unaffordable. In practice, we can perform rolling retraining once a month and adopt daily updates during the interval to learn new patterns timely. Our work is orthogonal to rolling retraining.

\subsection{Data Adaptation}
Apart from model adaptation, data adaptation is another mainstream approach to concept drift in time series forecasting, which is orthogonal to model adaptation. The goal of data adaptation is to normalize historical training data and future test data into a common data distribution~\cite{RevIN, Dish-TS, SAN}, reducing overfitting risks. Among them, RevIN~\cite{RevIN} is the most popular method that applies instance normalization to each time series lookback window and restores the statistics to the corresponding predictions, making each sample follow a similar distribution. Such normalization-based data adaptation methods mainly focus on the statistical changes in the mean and deviation of time series, while they overlook the distribution shifts in more complex temporal dependencies between time steps and spatial dependencies between variates~\cite{LIFT}. Meanwhile, the forecast model may underfit the normalized time series, as the removed statistics can serve as informative signals for prediction~\cite{NSTransformer}.

\section{Experimental Setup}
\begin{table}[h]
    \centering
    \small
    \caption{The statistics of five popular time series datasets.}
    \begin{tabular}{cccc}
    \toprule
Dataset & \# Sensors ($N$) & \# Time Steps & Frequency \\
    \midrule
ETTh2 & 7 & 17,420 & 1 hour \\
ETTm1 & 7 & 69,680 & 15 min \\
Weather & 21 & 52,696 & 10 min \\
ECL & 321 & 26,304 & 1 hour \\
Traffic & 862 & 17,544 & 1 hour \\  \bottomrule
    \end{tabular}
    \label{tab:dataset}
\end{table}
\subsection{Datasets}\label{sec:datasets}
We list descriptions of datasets used for experiments as follows.
\begin{itemize}[leftmargin=*]
    \item ETT~\cite{Informer} (Electricity Transformer Temperature) records seven features of electricity transformers from 2016 to 2018. ETTh1 is hourly collected and ETTm1 is 15-minutely collected. 
    \item ECL~\cite{DLinear} records the hourly electricity consumption of 321 clients from 2012 to 2014.
    \item Weather~\cite{Informer} records 21 features of weather, \textit{e.g.}, air temperature and humidity in 2020.
    \item Traffic~\cite{MTGNN} records the hourly road occupancy rates recorded by the sensors of San Francisco freeways from 2015 to 2016.
\end{itemize}


\subsection{Evaluation Settings}\label{sec:setting}
To the best of our knowledge, there is no open-source evaluation framework that considers the temporal gap in online learning for time series forecasting. Though Zhang \textit{et al.}~\cite{OneNet} noticed the feedback delay, they sidestep this issue by performing one prediction task every $H$ time steps and performing the next task until the ground truth is available. Concretely, they use $\mathbf{X}_{t-L:t}$ to forecast $\mathbf{X}_{t+1:t+H}$ and then use $\mathbf{X}_{t-L+H:t+H}$ to forecast $\mathbf{X}_{t+H+1:t+2H}$. In contrast, the common practice is to perform forecasting at each time step between $t$ and $t+H$, continually updating the previous predictions of $\mathbf{X}_{t-L+H:t+H}$. The reason is that short-term forecasting is typically easier than long-term forecasting. Therefore, in this work, we implement a more realistic evaluation framework for online learning where we perform forecasting at each time step, which remains consistent with the traditional evaluation setting.

As for the ETTm1 dataset, popular works (\textit{e.g.}, Informer~\cite{Informer}, PatchTST~\cite{PatchTST}, and iTransformer~\cite{iTransformer}) usually use 69,680 time steps of the dataset. In contrast, FSNet and OneNet only used a quarter of it, \textit{i.e.}, 17,420 steps, and the training set only contains 3484 samples. This setting is in favor of small models (\textit{e.g.}, TCN) but is challenging for the state-of-the-art time series forecast models. Therefore, we take all 69,680 time steps following PatchTST and iTransformer. As such, the ETTm1 dataset in our experiment has the greatest number of samples, enriching the diversity of our datasets.

\subsection{Implementation Details}\label{sec:details}
We conduct experiments on four Nvidia 4090 24GB GPUs. We ran each experiment 3 times with different random seeds and reported the average results. We follow the official implementation of the forecast models and the online learning methods, using their recommended model hyperparameters (\textit{e.g.}, the number of layers) and the Adam optimizer. In cases where default hyperparameter values are not provided, we conduct a grid search to find the optimal hyperparameters that yield the best performance. Following FSNet and OneNet, the lookback length $L$ for TCN is set to be $60$. For PatchTST and iTransformer, we search for the optimal lookback length and set $L$ to be $512$. 
We enhance TCN and FSNet by RevIN~\cite{RevIN} which is widely adopted by the state-of-the-art forecast models to reduce distribution shifts, lowering the forecasting loss in our experiments. As for the Optimal and Practical variants, we report the lowest errors when applying or not applying RevIN to FSNet and OneNet. Empirically, the Optimal variant without RevIN achieves better performance in some cases. 

For \modelns, we search the concept dimension $d_c$ in \{50, 100, 150, 200\} and the bottleneck dimension $r$ in \{24, 32, 48\}. The MLP used in $\mathcal E$ consists of a linear layer with weights in $\mathbb R^{(L+H)\times d_c}$, a GeLU activation function, and a linear year with weights in $\mathbb R^{d_c\times d_c}$, while the first linear layer of $\mathcal E'$ has weights in $\mathbb R^{L\times d_c}$.

\begin{table*}[t]
	\centering
	\caption{MAE results of two online learning strategies for time series forecasting. The horizon $H$ varies in 24, 48, and 96.  }\label{tab:mae gap}
  \resizebox{0.92\linewidth}{!}{
\begin{tabular}{cc|ccc|ccc|ccc|ccc|ccc}
\toprule
 & \multirow{2}{*}{\diagbox[width=2cm,height=0.85cm,innerrightsep=2pt,innerwidth=1.5cm]{}{\small \textbf{Dataset}\\$H$}} & \multicolumn{3}{c|}{\textbf{ETTh2}} & \multicolumn{3}{c|}{\textbf{ETTm1}} & \multicolumn{3}{c|}{\textbf{Weather}} & \multicolumn{3}{c|}{\textbf{ECL}} & \multicolumn{3}{c}{\textbf{Traffic}} \\
\multicolumn{1}{c}{\textbf{Model}} &  & \textbf{24} & \textbf{48} & \textbf{96} & \textbf{24} & \textbf{48} & \textbf{96} & \textbf{24} & \textbf{48} & \textbf{96} & \textbf{24} & \textbf{48} & \textbf{96} & \textbf{24} & \textbf{48} & \textbf{96} \\ \midrule
\multirow{3}{*}{FSNet} & \multicolumn{1}{c|}{Optimal} & 0.467 & 0.515 & \multicolumn{1}{l|}{0.596} & 0.472 & 0.538 & \multicolumn{1}{c|}{0.539} & 0.396 & 0.471 & \multicolumn{1}{c|}{0.511} & 0.397 & 0.434 & \multicolumn{1}{c|}{0.452} & 0.318 & 0.348 & 0.370 \\
 & \multicolumn{1}{c|}{Practical} & 0.705 & 0.829 & \multicolumn{1}{c|}{0.957} & 0.521 & 0.581 & \multicolumn{1}{c|}{0.609} & 0.406 & 0.558 & \multicolumn{1}{c|}{0.690} & 0.390 & 0.436 & \multicolumn{1}{c|}{0.474} & 0.325 & 0.355 & 0.384 \\
 & \multicolumn{1}{c|}{$\Delta_{\text{MAE}}$} & 51\% & 61\% & \multicolumn{1}{c|}{61\%} & 10\% & 8\% & \multicolumn{1}{c|}{13\%} & 2\% & 19\% & \multicolumn{1}{c|}{35\%} & -2\% & 0\% & \multicolumn{1}{c|}{5\%} & 2\% & 2\% & 4\% \\ \midrule
\multirow{3}{*}{OneNet} & \multicolumn{1}{c|}{Optimal} & 0.407 & 0.436 & \multicolumn{1}{l|}{0.545} & 0.272 & 0.275 & \multicolumn{1}{c|}{0.286} & 0.238 & 0.261 & \multicolumn{1}{c|}{0.276} & 0.333 & 0.348 & \multicolumn{1}{c|}{0.387} & 0.300 & 0.320 & 0.340 \\
 & \multicolumn{1}{c|}{Practical} & 0.728 & 0.902 & \multicolumn{1}{c|}{1.096} & 0.688 & 0.781 & \multicolumn{1}{c|}{0.678} & 0.456 & 0.591 & \multicolumn{1}{c|}{0.707} & 0.767 & 0.721 & \multicolumn{1}{c|}{0.505} & 0.319 & 0.366 & 0.393 \\
 & \multicolumn{1}{c|}{$\Delta_{\text{MAE}}$} & 79\% & 107\% & \multicolumn{1}{c|}{101\%} & 153\% & 184\% & \multicolumn{1}{c|}{137\%} & 92\% & 126\% & \multicolumn{1}{c|}{156\%} & 130\% & 107\% & \multicolumn{1}{c|}{31\%} & 7\% & 14\% & 16\% \\ \midrule
\multirow{3}{*}{PatchTST} & \multicolumn{1}{c|}{Optimal} & 0.581 & 0.649 & \multicolumn{1}{c|}{0.773} & 0.336 & 0.334 & \multicolumn{1}{c|}{0.335} & 0.325 & 0.336 & \multicolumn{1}{c|}{0.335} & 0.295 & 0.316 & \multicolumn{1}{c|}{0.342} & 0.287 & 0.274 & 0.277 \\
 & \multicolumn{1}{c|}{Practice} & 0.611 & 0.718 & \multicolumn{1}{c|}{0.864} & 0.407 & 0.474 & \multicolumn{1}{c|}{0.509} & 0.369 & 0.482 & \multicolumn{1}{c|}{0.592} & 0.294 & 0.315 & \multicolumn{1}{c|}{0.341} & 0.276 & 0.267 & 0.278 \\
 & \multicolumn{1}{c|}{$\Delta_{\text{MAE}}$} & 5\% & 11\% & \multicolumn{1}{c|}{12\%} & 21\% & 42\% & \multicolumn{1}{c|}{52\%} & 14\% & 43\% & \multicolumn{1}{c|}{77\%} & 0\% & 0\% & \multicolumn{1}{c|}{0\%} & -4\% & -2\% & 0\% \\ 
 \bottomrule
  \end{tabular}
  }
\end{table*}

\section{Pipeline of \model}
\begin{figure}[h]
    \centering
    \includegraphics[width=\linewidth]{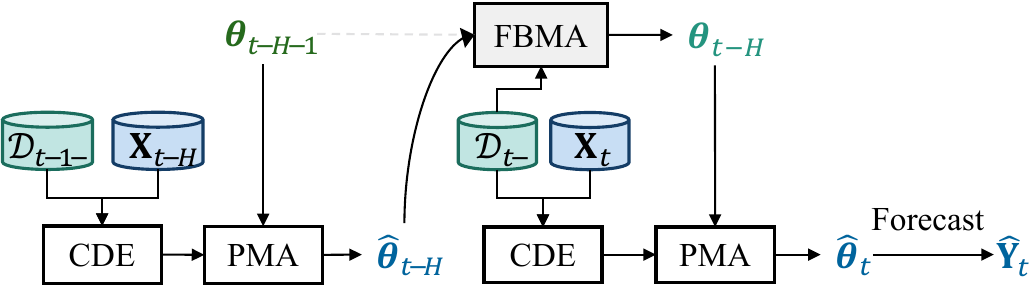}
    \caption{Pipeline of \model at each online time $t$. CDE: concept drift estimation; PMA: proactive model adaptation.}
    \label{fig:pipeline}
\end{figure}
Fig.~\ref{fig:pipeline} depicts the pipeline of \model during the online phase, which is a bit different from the steps introduced in Sec.~\ref{sec:overview}. Given model parameters $\boldsymbol{\theta}_{t-H-1}$ that have been updated with previous training data $\mathcal D_{(t-1)-}$, our model adapter additionally estimates the concept drift between $\mathcal D_{(t-1)-}$ and $\mathcal D_{t-}$ and adapts the model to $\mathcal D_{t-}$, yielding forecast feedback for updates. The reason behind it is that we jointly train the forecast model and the model adapter (see Sec.~\ref{sec:training}), while our training algorithm does not enforce the forecast model to perform well alone. Consequently, the forecast model without $\Delta\boldsymbol{\theta}$ could forget some predictive skills, while the parameter adjustments $\Delta\boldsymbol{\theta}$ from the model adapter may incorporate the skills forgotten by the model. Thus it is necessary for the inference process to keep consistent with the training phase, \textit{i.e.}, we always make predictions by integrating the forecast model and our model adapter as a whole. Nevertheless, in the feedback-based adaptation step, we only finetune the forecast model but keep the model adapter's parameters frozen. Otherwise, the model adapter may overfit the patterns of the one-step concept drift between $\mathcal D_{(t-1)-}$ and $\mathcal D_{t-}$, losing generalization ability.

\section{Further Experimental Results}
\subsection{Performance Gap in terms of MAE}\label{sec:gap-mae}
Table~\ref{tab:mae gap} shows the MAE errors of two variants of online learning introduced in Sec.~\ref{sec:analysis}. Likewise, we define $\Delta_{\text{MAE}}$ as the performance gap \textit{w.r.t.} MAE. We can observe a general trend similar to Table~\ref{tab:mse gap}, where a longer horizon can enlarge the performance gap between the Optimal and the Practical variants. The ECL and Traffic datasets do not have significant concept drift between each test sample and its preceding training sample. By contrast, $\Delta_{\text{MSE}}$ is more significant than $\Delta_{\text{MAE}}$. We conjecture that online learning mainly benefits the forecast models in some rare or extreme cases of time series evolution where the frozen model's predictions deviate far from the ground truth and result in large MSE. 

\subsection{Performance with Partial Ground Truth}~\label{sec:partial}
\begin{table}[th]
\centering
\caption{MSE when using $\mathbf Y_{t-H}$ and partial ground-truth.}\label{tab:mse partial}
\setlength\tabcolsep{2pt}
  \resizebox{\linewidth}{!}{
\begin{tabular}{cc|ccc|ccc|ccc}
\toprule
 & \multirow{2}{*}{\diagbox[width=2cm,height=0.85cm,innerrightsep=2pt,innerwidth=1.5cm]{\textbf{\small Method}}{\small \textbf{Dataset}\\$H$}} & \multicolumn{3}{c|}{\textbf{ETTh2}} & \multicolumn{3}{c|}{\textbf{ETTm1}} & \multicolumn{3}{c}{\textbf{Weather}} \\
\multicolumn{1}{c}{\textbf{Model}} &  & \textbf{24} & \textbf{48} & \textbf{96} & \textbf{24} & \textbf{48} & \textbf{96} & \textbf{24} & \textbf{48} & \textbf{96}\\ \midrule
\multirow{2}{*}{PatchTST} & GD & 1.833 & 3.144 & 5.429 & 0.455 & 0.591 & 0.673 & 0.734 & 0.977 & 1.261 \\
 & \modelns & \textbf{1.703} & \textbf{3.070} & \textbf{5.299} & \textbf{0.420} & \textbf{0.570} & \textbf{0.650} & \textbf{0.719} & \textbf{0.967} & \textbf{1.256} \\\midrule
 \multirow{2}{*}{iTransfo.} & GD & 2.463 & 3.759 & 6.152 & 0.459 & 0.602 & 0.692 & 0.851 & 1.131 & 1.406 \\
 & \modelns & \textbf{2.228} & \textbf{3.571} & \textbf{5.868} & \textbf{0.426} & \textbf{0.569} & \textbf{0.659} & \textbf{0.734} & \textbf{1.001} & \textbf{1.299}\\
 \bottomrule
\end{tabular}
}
\end{table}

In this section, we introduce variants that update the model with feedback on $\mathbf Y_{t-H}$ and partial ground truth $\{\tilde{\mathbf Y}_{t-i}\}_{i=1}^{H-1}$, where ${\tilde{\mathbf Y}_{t-i}=[\mathbf v_{t-i},\cdots, \mathbf v_{t-1}]}$. Since each update involves $H$ training samples, the GPU memory overhead becomes about $H$ times than one involving $\mathbf Y_{t-H}$ only. We leave out results on ECL and Traffic datasets, where the GPU memory overhead exceeds the limit of our Nvidia 4090 GPUs (24GB).
As shown in Table~\ref{tab:mse partial}, \model consistently outperforms the online gradient descent method under this setting. Compared with Table~\ref{tab:main_result_mse}, using both $\mathbf Y_{t-H}$
 and partial ground truth for feedback-based adaptation can improve the performance, while the improvement is insignificant. This indicates that samples with partial ground truth cannot reveal the concept of the test sample, necessitating proactive model adaptation. Though the partial ground truth can be beneficial to our method, the finetuning cost scales up with $H$ (e.g., up to either 96$\times$ latency or 96$\times$ GPU memory). Thus, it would be much more efficient if using $\mathbf Y_{t-H}$ only. Notably, our proposed method using $\mathbf Y_{t-H}$ only can still outperform GD using $H$ training samples in most cases. 
. 

\balance

\end{document}